\documentclass[11pt]{article}

\usepackage[preprint]{acl}

\usepackage{times}
\usepackage{latexsym}

\usepackage[T1]{fontenc}

\usepackage[utf8]{inputenc}

\usepackage{microtype}

\usepackage{inconsolata}

\usepackage{graphicx}
\usepackage{amsmath}
\usepackage{amssymb}
\usepackage{booktabs}

\usepackage{bm}
\usepackage{float}
\usepackage{tcolorbox}
\usepackage{multirow}
\usepackage{colortbl}
\usepackage{tabulary}
\usepackage{etoolbox}
\usepackage{pifont}
\usepackage{makecell}
\usepackage{bbm}
\usepackage[percent]{overpic}
\usepackage{adjustbox}
\usepackage{rotating}
\usepackage{tablefootnote}
\definecolor{LightCyan}{rgb}{0.8,1,1}
\definecolor{LightGreen}{rgb}{0.6,0.9,0.7}
\definecolor{royalblue(traditional)}{rgb}{0.0, 0.14, 0.4}
\definecolor{royalblue(web)}{rgb}{0.25, 0.41, 0.88}
\definecolor{darkgreen}{rgb}{0.0, 0.5, 0.0} 
\definecolor{darkred}{rgb}{0.5, 0.0, 0.0}
\definecolor{darkblue}{rgb}{0.0, 0.0, 0.8}
\usepackage{subcaption}

\title{BioVLM: Routing Prompts, Not Parameters, for Cross-Modality Generalization in Biomedical VLMs}

\author{
Mainak Singha$^{1}$, Tanisha Gupta$^{2}$, Ankit Jha$^{3}$, Muhammad Haris Khan$^{4}$,\\
\textbf{Sayantani Ghosh}$^{5}$, \textbf{Biplab Banerjee}$^{6}$\\
$^{1}$University of Trento, Italy \quad
$^{2}$Carnegie Mellon University, USA \quad
$^{3}$LNMIIT Jaipur, India\\
$^{4}$MBZUAI, UAE \quad
$^{5}$Sunandan Divatia School of Science, Mumbai, India \quad
$^{6}$IIT Bombay, India
}

\begin{document}
\maketitle
\begin{abstract}

Pretrained biomedical vision-language models (VLMs) such as BioMedCLIP perform well on average but often degrade on challenging modalities where inter-class margins are small and acquisition-specific variations are pronounced, especially under few-shot supervision and when modality priors differ from pretraining corpora substantially. We propose BioVLM, a prompt-learning framework that improves cross-domain generalization without extensive backbone fine-tuning. BioVLM learns a diverse prompt bank and introduces dynamic prompt selection: for each input, it selects the most discriminative prompts via a low-entropy criterion on the predictive distribution, effectively coupling sparse few-shot evidence with rich LLM semantic priors. To strengthen this coupling, we distill high-confidence LLM-derived attributes and enforce robust knowledge transfer through strong/weak augmentation consistency. At test time, BioVLM adapts by choosing modality-appropriate prompts, enabling transfer to unseen categories and domains, while keeping training lightweight and inference efficient. On 11 MedMNIST+ 2D datasets, BioVLM achieves new state of the art across three distinct generalization settings. Codes are available at \url{https://github.com/mainaksingha01/BioVLM}. 
\end{abstract}

\begin{figure}[!htbp] 
    \centering
    \includegraphics[width=\columnwidth]{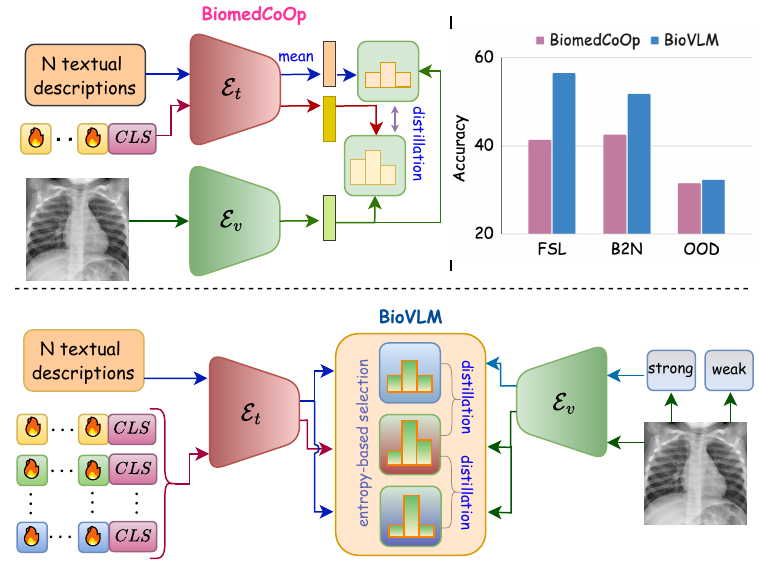} 
    \vspace{-0.6cm}
    \caption{\textbf{Overview of our proposed BioVLM.} It selects high-confidence, optimal prompts using an entropy-based selection strategy and synergistically distills few-shot task semantics with the rich prior generic knowledge. BioVLM significantly outperforms the SOTA baseline, BioMedCoOp \cite{biomedcoop}, across three distinct generalization settings.}
    \label{fig:teaser}
    \vspace{-0.7cm}
\end{figure}

\section{Introduction}
\label{sec:intro}
Foundation models pretrained on web-scale image-text pairs, such as CLIP \cite{clip} and ALIGN \cite{align}, have reshaped vision-language learning by enabling strong zero-shot transfer via contrastive cross-modal alignment. Yet, their performance often collapses when deployed ``as-is'' in biomedicine. The root cause is a persistent domain mismatch: biomedical images (e.g., MRI, ultrasound, dermatoscopy) exhibit modality-dependent signal formation, characteristic textures, and acquisition artifacts that differ fundamentally from natural-image statistics. In addition, biomedical supervision is textually fragile reports and labels are sparse, jargon-heavy, and require domain expertise, making the language side noisier and less visually grounded than web captions, which challenges models trained on general-domain text.

To reduce this mismatch, biomedical VLMs such as BioMedCLIP \cite{BioMedCLIP}, PubMedCLIP \cite{pubmedclip}, and BioViL \cite{biovil} pretrain on in-domain image-text corpora, learning clinically relevant cross-modal representations. These models have enabled progress in downstream settings including pathology \cite{plip, quilt} and radiology \cite{medclip, radiology}, where fine-grained visual cues and structured medical semantics are critical. Consequently, they provide strong frozen backbones for medical AI, capturing complex morphology and modality-specific patterns that general VLMs typically miss.

Despite these advances, a key research gap remains: even specialized biomedical VLMs can be unreliable in zero-shot transfer to unseen conditions and modalities, especially when those distributions are underrepresented during pretraining and when clinical text contains ambiguity or semantic noise. Prompt learning is an attractive, parameter-efficient way to adapt frozen backbones, but existing methods remain limited. Approaches from CoOp \cite{coop} to BioMedCoOp \cite{biomedcoop} typically optimize a single prompt (or a small ensemble), which is insufficient to model the diversity of biomedical appearance across modalities and scanners. Moreover, BioMedCoOp integrates LLM knowledge \cite{gpt4} by averaging and distilling multiple descriptions into one vector, which can propagate redundant or weakly grounded attributes and place disproportionate burden on a single prompt to generalize in unseen cases. In parallel, TPT \cite{tpt} selects confident augmented views for test-time adaptation; however, for biomedical images, common augmentations (e.g., cropping/zooming) can perturb subtle diagnostic evidence and thereby bias confidence-based selection.

To address these limitations, we propose \textbf{BioVLM}, a prompt-learning framework that improves the generalization of frozen biomedical VLMs across heterogeneous medical imaging tasks (Fig.~\ref{fig:teaser}). Built on BioMedCLIP, BioVLM introduces \emph{dynamic prompt engineering} with two explicit mechanisms. First, we maintain a \emph{diverse bank of learnable prompts} and perform \emph{low-entropy prompt selection} to choose the most informative prompts per input, rather than committing to a single prompt. Second, unlike TPT \cite{tpt}, we select \emph{high-confidence textual features}: (i) we retrieve and distill the LLM descriptions that are most aligned with the input visual features (jointly under weak/strong augmentations), and (ii) we select the subset of prompts whose textual embeddings best match the input semantics in the shared vision-language space. This design filters noisy LLM supervision, reduces redundancy, and emphasizes the most discriminative cross-modal alignments. At inference, BioVLM adapts by selecting highly aligned prompts for the target modality, enabling robust transfer to unseen categories and modalities without requiring LLM plug-ins. Our primary contributions are summarized as follows:

\noindent - \textbf{A novel prompt learning framework, BioVLM,} specifically engineered to enhance the generalization of frozen biomedical VLMs across complex and unseen medical imaging modalities.

\noindent - \textbf{A low-entropy prompt selection strategy} that dynamically identifies the most discriminative prompts from a diverse learnable pool, effectively aligning few-shot task knowledge with LLM priors.

\noindent - \textbf{A robust cross-modal alignment technique} that distills high-confidence attributes from an LLM and leverages a dual strong/weak image augmentation strategy to improve knowledge transfer.

\noindent - \textbf{Demonstrated state-of-the-art performance} through extensive experiments on 11 datasets from the MedMNIST+ benchmark, where BioVLM consistently outperforms existing baselines in three distinct generalization settings (Fig.~\ref{fig:teaser}).

\section{Related Works}
\label{sec:related_works}

\noindent \textbf{(i) Biomedical Vision-Language Models:}
The success of generic VLMs such as CLIP~\cite{clip} has motivated biomedical counterparts to reduce the mismatch between natural and clinical data. Models such as BioViL~\cite{biovil}, PubMedCLIP~\cite{pubmedclip}, and BioMedCLIP~\cite{BioMedCLIP} pretrain on large-scale in-domain image-text corpora and adopt domain-aware choices (e.g., radiology-initialized language encoders, medical vocabularies, and clinical contrastive objectives) to improve cross-modal alignment. Beyond general biomedical VLMs, modality- or task-specific foundations target pathology~\cite{plip, quilt}, chest X-ray~\cite{medclip}, radiology~\cite{radiology}, and retinal imaging~\cite{flairretina}, often using multi-modal co-attention and masking of domain-relevant terms~\cite{gan2022vision} to boost zero-shot retrieval and recognition. Nevertheless, learning fine-grained, disease-specific semantics that transfer across modalities, scanners, and institutions remains challenging; as noted in~\cite{zhou2022generalized}, no single pretrained model covers all downstream needs.

Even specialized biomedical VLMs can be brittle under modality shift and long-tail conditions, motivating lightweight task/modality adaptation. BioVLM performs inference-time prompt routing to adapt a frozen biomedical VLM to the target modality without extensive fine-tuning.

\noindent \textbf{(ii) Prompt Learning:}
Prompt learning adapts frozen VLMs by optimizing prompts rather than updating backbone parameters. In the general domain, CoOp~\cite{coop} learns continuous textual prompts for few-shot classification, and CoCoOp~\cite{cocoop} conditions prompts on image features for better generalization. Later work injects structured priors (KgCoOp~\cite{kgcoop}) and semantic regularization (ProGrad~\cite{prograd}). Prompting has expanded to multi-modal prompts (MaPLe~\cite{maple}, PromptSRC~\cite{promptsrc}), deeper prompt stacks (TCP~\cite{tcp}), and visual prompting (VPT~\cite{vpt}), and has been explored in domain adaptation~\cite{clipoint3d, cosmo}, open-world segmentation~\cite{promptseg, openseg}, and federated learning~\cite{fedclip, fedmvp}. In biomedicine, BioMedCoOp~\cite{biomedcoop} builds on BioMedCLIP using LLM-generated descriptions with statistics-based selection, but prompting can also expose backdoor vulnerabilities in medical settings~\cite{baple, fedmedbackdoor}.

Most prompt methods still optimize a single prompt (or small ensemble) and lack principled filtering of noisy supervision or dynamic selection under modality shift. BioVLM maintains a diverse prompt bank and uses low-entropy selection with LLM-attribute distillation to denoise and route the best prompts per input and modality.

\section{Proposed Methodology}
\label{sec:methodology}
\subsection{Problem Formulation}
We formalize the task as follows. Let $\mathcal{D}_s=\{(\mathbf{x}_i,y_i)\}_{i=1}^{n}$ denote the labeled source dataset, where $\mathbf{x}_i\in\mathcal{X}$ and $y_i\in\mathcal{C}_{\mathrm{train}}$, with $|\mathcal{C}_{\mathrm{train}}|=N_c$. We learn a classifier on $\mathcal{D}_s$ using supervision from $\mathcal{C}_{\mathrm{train}}$. At inference, given a target dataset $\mathcal{D}_t$, for each query image $\mathbf{x}\in\mathcal{D}_t$ we compute class-wise similarity scores $\{s(\mathbf{x},c)\}_{c\in\mathcal{C}_{\text{test}
}}$ in a shared embedding space and predict
$\hat{y}(\mathbf{x})=\arg\max_{c\in\mathcal{C}_{\text{test}}} s(\mathbf{x},c)$.
Following standard biomedical evaluation protocols~\cite{medmnist, BioMedCLIP}, we report results under three generalization settings:

\noindent \textbf{- Few-Shot Learning:}
Evaluates sample efficiency when only $\mathcal{K}$ labeled instances per class are available for training, with identical label spaces at train and test, i.e., $\mathcal{C}_{\mathrm{train}}=\mathcal{C}_{\mathrm{test}}$.

\noindent \textbf{- Base-to-New Generalization:}
Assesses transfer to novel categories within the same modality by using disjoint label sets,
$\mathcal{C}_{\mathrm{train}}\cap\mathcal{C}_{\mathrm{test}}=\emptyset$, where $\mathcal{C}_{\mathrm{test}}$ contains unseen classes.

\noindent \textbf{- Out-of-Distribution (OOD) Generalization:}
Tests robustness to domain shift where the target distribution differs from the source, i.e.,
$P_t(\mathbf{x},y)\neq P_s(\mathbf{x},y)$, while the label space may be shared or partially overlapping depending on the protocols.

\begin{figure}[t]
\scalebox{0.95}{
    \centering
    \vspace{-1.1cm}
    \includegraphics[width=\columnwidth]
    {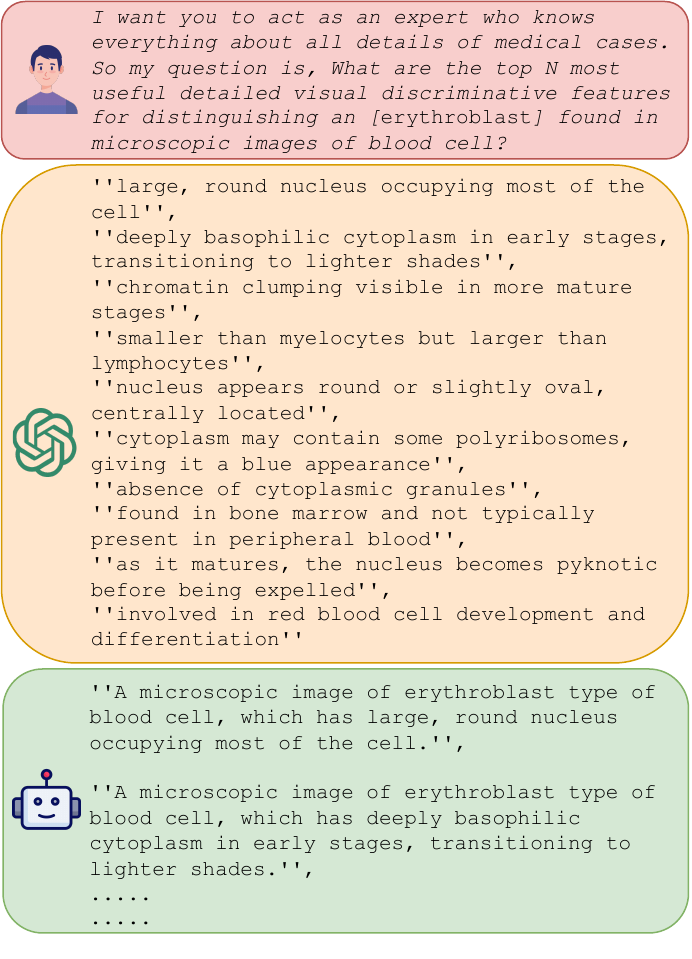}}
    \vspace{-0.5cm}
    \caption{\textbf{LLM attribute generation.} To derive high-level clinical knowledge representations, we follow a three-stage approach. First (\textbf{top box}), an instructional query prompt is provided to a Large Language Model (LLM). In response, the LLM generates detailed visual and clinical descriptions (\textbf{middle box}). Finally (\textbf{bottom box}), we construct highly contextualized textual prompts by combining a modality-specific prefix template with the LLM-generated attributes.}
    \vspace{-0.45cm}
    \label{fig:llm}
\end{figure}

\subsection{The BioVLM Framework: From Static to Adaptive Prompting}
\label{sec:method}

Our BioVLM framework operates on a frozen CLIP backbone, which includes a vision encoder $\mathcal{E}_v$ and a text encoder $\mathcal{E}_t$. For any given image $\mathbf{x}$, a visual embedding is first extracted as $\mathbf{V} = \mathcal{E}_{v}(\mathbf{x})$. The core of our method lies in how we engineer the corresponding textual embeddings for robust cross-modal alignment.

We begin with a prompt structure inspired by CoOp~\cite{coop}, where a class prompt $\mathbf{t}_i$ is formed by learnable context vectors $\{\mathbf{p}_1, \dots, \mathbf{p}_M\}$ prepended to a class token embedding $\mathbf{c}_i$. However, we posit that a single, generic prompt is insufficient for the diverse and complex nature of medical imaging. Our first key contribution is to create a \textbf{diverse prompt pool}. Using an LLM, we generate multiple rich, descriptive attributes for the symptomatic features of each medical condition. For each attribute, a unique learnable prompt is instantiated, forming a comprehensive set of textual descriptions per class. 

With this diverse pool, our second contribution is a \textbf{dynamic prompt selection mechanism}. For a given image, we must identify the most suitable prompt. We achieve this with a confidence-based strategy that selects the prompt yielding the lowest entropy over the final classification probabilities. Low entropy signifies a high-confidence, unambiguous prediction, indicating a strong alignment between the image's visual evidence and the prompt's semantic meaning.

Finally, to ensure this selection process is robust and generalizes to unseen data, we introduce a \textbf{consistency-based distillation} step. We apply both strong and weak augmentations to the input image and require our model to make consistent, high-confidence predictions across these variations. By optimizing for a prompt that performs well across all views, we effectively distill robust visual features and guide the model to learn semantically meaningful representations that are invariant to superficial noise, shown in Table \ref{tab:loss}. The stages are detailed in the following.

\noindent\textbf{(a) LLM attribute generation:} 
We generate descriptive attributes for each class using GPT-4o~\cite{gpt4o}, queried with a structured instructional prompt adapted from~\cite{desc} (Figure~\ref{fig:llm}). To construct contextualized textual prompts, we combine a modality-specific prefix with the LLM-generated attributes. The prefix can be generic or tailored to the imaging modality (e.g., \texttt{``a microscopic image of a [class name] kidney cortex cell''}). This prefix is concatenated with an attribute using connectors such as \texttt{``which is''} or \texttt{``which has''}, yielding modality-aware and semantically rich composite prompts. The third row of Figure~\ref{fig:llm} provides a complete example of such a prompt generated for the class \texttt{``erythroblast''}.

\begin{figure*}[htbp] 
    \centering
    \includegraphics[width=\textwidth]{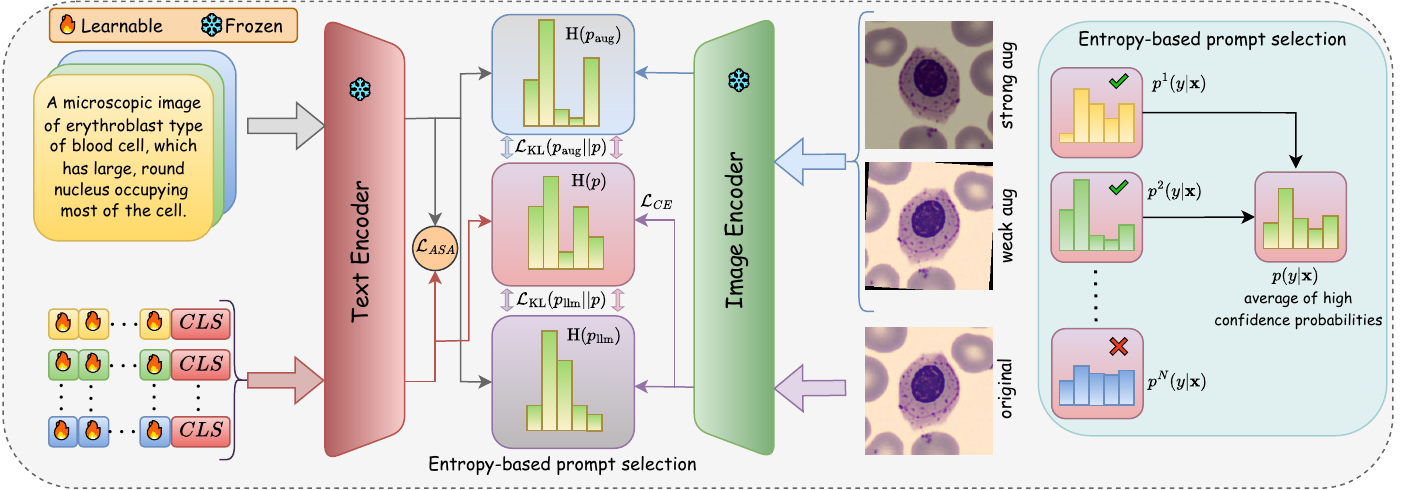}
    \vspace{-0.5cm}
\caption{\textbf{Model architecture of BioVLM.} The proposed framework enhances the generalization capability of the pretrained BioMedCLIP by integrating learnable prompts with LLM-derived attributes processed through a frozen text encoder. An entropy-based selection strategy identifies the most discriminative prompts. Original, weakly augmented, and strongly augmented images are encoded using a frozen image encoder. The model is trained using cross-entropy loss ($\mathcal{L}_{CE}$), low-entropy regularization losses, KL-divergence distillation losses ($\mathcal{L}_{KL}$), and an alignment loss ($\mathcal{L}_{ASA}$). The final prediction is obtained by averaging high-confidence outputs.}
\vspace{-0.3cm}
\label{fig:main_fig}
\end{figure*}

\noindent\textbf{(b) Diverse prompt tuning:} 
While a single learnable prompt vector can effectively distill knowledge from multiple attributes and learn image-text alignment~\cite{coop, biomedcoop}, this approach has inherent limitations. A single prompt tends to learn an averaged or blended representation of all attributes, which can compromise its specificity and reduce generalization, particularly when encountering unseen classes with unfamiliar feature distributions.

To overcome this, we propose a \textbf{multi-prompt framework}. Instead of a single vector, we initialize a \textbf{diverse prompt pool} for each class. For class $i$, we define $N$ distinct prompts:
\[
\mathbf{t}_i = \{\mathbf{t}_i^j\}_{j=1}^{N}, \quad 
\mathbf{t}_i^j = \{\mathbf{p}^{j}_1, \dots, \mathbf{p}^{j}_M, \mathbf{c}_i\},
\]
where each prompt combines learnable context vectors with the class token. For a given image, this pool produces $N$ textual features, enabling multiple vision-language alignments, but raising the question: \textit{how to select the most reliable prompt for accurate prediction?}

\noindent \textbf{(c) Entropy-based prompt selection:}
To select the most discriminative prompt from the pool, we use an entropy-based mechanism: prompts that align well with an image’s visual features yield low-entropy (high-confidence) class distributions. For a given image with visual features $\mathbf{V}$, the probability distribution from prompt variant $j$ is computed as the softmax over cosine similarities:
\vspace{-0.1cm}
\begin{equation}
    p^j(y|\mathbf{x}) = \frac{\exp(\text{cos}(\mathbf{V}, \mathbf{T}_y^j)/\beta)}{\sum_{i=1}^{K} \exp(\text{cos}(\mathbf{V}, \mathbf{T}_i^j)/\beta)}
    \label{eq:prob}
\end{equation}
Here, $\mathbf{T}_i^j = \mathcal{E}_t(\mathbf{t}_i^j)$  denotes the feature vector of the $j$-th prompt for class $i$, $K$ is the total number of classes, and $\beta$ is a temperature hyperparameter. The self-entropy of this distribution, measuring its uncertainty, is defined as:
\vspace{-0.1cm}
\begin{equation} 
    \mathrm{H}(p^j) = - \sum_{i=1}^{K} p^j(y_i|\mathbf{x}) \log p^j(y_i|\mathbf{x}).
    \label{eq:entropy}
\end{equation}

Rather than using a fixed entropy threshold, which may not adapt well across different images or datasets, we employ a more robust percentile-based strategy. We calculate the entropy $\mathrm{H}(p^j)$ for all $N$ prompts and identify the entropy value at a given cutoff percentile, $\rho$. Let this threshold be $\tau_{\rho}$. We then select only the prompts whose entropy is below this threshold. The final class prediction probability is the normalized average of the probabilities from these high-confidence prompts:
\begin{equation}
    p(y|\mathbf{x}) = \frac{\sum_{j=1}^{N} \mathbbm{1}[\mathrm{H}(p^j) \leq \tau_{\rho}] \cdot p^j(y|\mathbf{x})}{\sum_{j=1}^{N} \mathbbm{1}[\mathrm{H}(p^j) \leq \tau_{\rho}]},
    \label{eq:final_prob}
\end{equation}
where $\mathbbm{1}[\cdot]$ is the indicator function. This ensures that only reliable prompts influence the final decision, improving robustness and generalization.

\subsection{Training \& Inference}
For training, we employ a composite loss to align learnable prompts with LLM-derived semantics. We extract frozen text features for the $N$ attributes of each class $i$ using the text encoder $\mathcal{E}_t$, denoted as $\mathbf{T}^{\text{llm}}_{i}=\{\mathbf{T}^{\text{llm},(j)}_{i}\}_{j=1}^{N}$, which serve as semantic targets for prompt learning.

 \textbf{- Attribute-Semantic alignment ($\mathcal{L}_{ASA}$):}
A core component of our method is the one-to-one mapping between the $N$ learnable prompts and the $N$ LLM-generated attributes for each class. This correspondence ensures that each prompt specializes in distilling knowledge from a specific attribute, preventing the feature averaging that can dilute discriminative information. To enforce this, our Attribute-Semantic Alignment loss maximizes the cosine similarity between each learnable prompt's features ($\mathbf{T}_i^j$) and its corresponding LLM attribute's features ($\mathbf{T}_i^{\text{llm}, (j)}$). This is equivalent to minimizing the negative similarity:
\begin{equation}
    \mathcal{L}_{ASA} = - \sum_{i=1}^{K} \sum_{j=1}^{N} \frac{\mathbf{T}_i^j \cdot \mathbf{T}_i^{\text{llm}, (j)}}{\|\mathbf{T}_i^j\| \cdot \|\mathbf{T}_i^{\text{llm}, (j)}\|}.
    \label{eq:L_asa}
\end{equation}

\textbf{- Low-Entropy regularization ($\mathcal{L}_{LER}$):}
To improve the model's prediction confidence and discriminative ability, we apply an entropy minimization loss to three key probability distributions.\\
\noindent \textbf{(a) Student Confidence:} We minimize the entropy of the final prediction $p(y|\mathbf{x})$ (from Eq.~\ref{eq:final_prob}) to encourage confident and unambiguous classifications.

\noindent \textbf{(b) Teacher Confidence:} Since not all LLM-generated attributes are equally discriminative, we additionally enforce confidence on the teacher distributions. 
Specifically, $p_{\mathrm{llm}}(y\mid\mathbf{x})$ is computed by matching the image embedding with frozen LLM-derived attribute embeddings, while the augmentation-based distribution is defined as
$p_{\mathrm{aug}}(y\mid\mathbf{x})=\tfrac{1}{2}\big(p(y\mid\mathbf{x}^{w})+p(y\mid\mathbf{x}^{s})\big)$.
We minimize the entropy of both $p_{\mathrm{llm}}$ and $p_{\mathrm{aug}}$ to encourage confident and reliable teacher supervision.

The combined low-entropy regularization is formulated in Eq. \ref{eq:L_ler} with $\mathrm{H}(\cdot)$ is the entropy function defined in Eq.~\ref{eq:entropy}.
\begin{equation}
    \mathcal{L}_{LER} = \mathrm{H}(p) + \mathrm{H}(p_{\text{llm}}) + \mathrm{H}(p_{\text{aug}}),
    \label{eq:L_ler}
\end{equation}

\textbf{- Cross-Modal Knowledge Distillation ($\mathcal{L}_{CMD}$):}
We transfer semantic knowledge from LLM-derived distributions (teachers) to the learnable prompt-based model (student) by minimizing the Kullback-Leibler (KL) divergence between student logits $p$ and both the LLM-based teacher $p_{\text{llm}}$ and the augmented-view teacher $p_{\text{aug}}$. This encourages the student to match the teachers’ confident and robust predictions:
\begin{equation}
    \mathcal{L}_{CMD} = \mathcal{L}_{\text{KL}}(p_{\text{llm}} || p) + \mathcal{L}_{\text{KL}}(p_{\text{aug}} || p).
    \label{eq:L_cmd}
\end{equation}

\noindent \textbf{Overall Training Objective:}
The final training objective combines the standard cross-entropy classification loss ($\mathcal{L}_{CE}$) with our proposed regularization and distillation terms, weighted by hyperparameters $\lambda_1, \lambda_2, \lambda_3$:
\begin{equation}
    \mathcal{L} = \mathcal{L}_{CE} + \lambda_{1}\mathcal{L}_{ASA} + \lambda_{2}\mathcal{L}_{LER} + \lambda_{3}\mathcal{L}_{CMD}.
    \label{eq:total_loss}
\end{equation}

From a \textbf{\textit{generalization theory perspective}}, our combined loss $\mathcal{L}$ is designed to minimize the true risk, $\mathcal{R}_{true}(f)$, by tackling both terms in the standard learning bound: $\mathcal{R}_{true}(f) \le \mathcal{R}_{emp}(f) + \Omega(\mathcal{H})$. While the cross-entropy loss ($\mathcal{L}_{CE}$) directly minimizes the empirical risk $\mathcal{R}_{emp}(f)$, the remaining objectives synergistically regularize the model to reduce the complexity term $\Omega(\mathcal{H})$. The attribute-semantic alignment loss ($\mathcal{L}_{ASA}$) prunes the hypothesis space $\mathcal{H}$ to a lower-complexity, semantically meaningful subspace defined by the LLM priors. Concurrently, the low-entropy loss ($\mathcal{L}_{LER}$) acts as a large-margin regularizer, pushing for more decisive predictions, which is known to improve generalization bounds. Finally, the knowledge distillation loss ($\mathcal{L}_{CMD}$) serves as a functional regularizer, encouraging the student model to find smoother functions that occupy flatter minima in the loss landscape by mimicking a stable teacher. This multifaceted regularization strategy ensures that the model learns a function that is not just accurate on training data but also simple, decisive, and smooth, all hallmarks of a model with a tight generalization bound and robust performance on unseen data.

\noindent \textbf{Inference:} During inference, our method is highly efficient, relying only on optimized learnable prompts and entropy-based selection. LLM-based attribute generation and image augmentations are used exclusively during training, substantially reducing test-time computational overhead.

\begin{table*}[!ht]
    \centering
    \caption{\textbf{Comparison of methods on the few-shot learning task of MedMNIST+ benchmark.}}
    \vspace{-0.2cm}
    \scalebox{0.6}{
    \begin{tabular}{l|ccccccccc|cc}
    \toprule
         &\multicolumn{1}{c}{BioMedCLIP} &\multicolumn{1}{c}{CoOp} &\multicolumn{1}{c}{CoCoOp} &\multicolumn{1}{c}{KgCoOp} &\multicolumn{1}{c}{MaPLe} &\multicolumn{1}{c}{ProGrad} &\multicolumn{1}{c}{PromptSRC} &\multicolumn{1}{c}{TCP} &\multicolumn{1}{c}{BiomedCoOp}& BioVLM &$\Delta$ \\
        
        \multirow{-2}{*}{\textbf{Dataset}} &NEJM AI'25 &IJCV'22 &CVPR'22 &CVPR'23 &CVPR'23 &ICCV'23 &ICCV'23 &CVPR'24 &CVPR'25 & (Ours) & (in \%) \\ 
        \midrule

        \textbf{Train Params} & 0 &3K &44K &3K &5.3M &3K &69K &0.5M &3K &\cellcolor[gray]{0.9}{30K} &\cellcolor[gray]{0.9}{-} \\

        \midrule

        \textbf{PathMNIST}	&43.30	&76.93	&75.92	&76.61 &75.50	&60.26 &76.42	&76.21	&72.59	&\cellcolor[gray]{0.9}\textbf{81.56} &\cellcolor[gray]{0.9}\textcolor{darkgreen}{+4.63} \\ 
        \textbf{DermaMNIST}	&36.21	&28.78	&40.55	&30.09 &37.53	&35.88 &39.80	&39.97	&30.02	&\cellcolor[gray]{0.9}\textbf{45.27} &\cellcolor[gray]{0.9}\textcolor{darkgreen}{+4.72} \\ 
        \textbf{OCTMNIST}	&36.80	&59.20	&60.47	&57.67 &60.76	&47.03 &59.35	&59.92	&54.77	&\cellcolor[gray]{0.9}\textbf{62.57} &\cellcolor[gray]{0.9}\textcolor{darkgreen}{+1.81} \\ 
        \textbf{PneumoniaMNIST}	&59.46	&72.70	&76.01	&73.24 &74.81	&57.64 &75.06	&72.48 	&70.51	&\cellcolor[gray]{0.9}\textbf{79.65}& \cellcolor[gray]{0.9}\textcolor{darkgreen}{+3.64} \\ 
        \textbf{RetinaMNIST}	&35.25	&31.58	&\textbf{40.92}	&30.83 &38.45	&29.17 &38.95	&39.68 &34.50	&\cellcolor[gray]{0.9}40.08 &\cellcolor[gray]{0.9}\textcolor{red}{-0.84} \\ 
        \textbf{BreastMNIST} &33.33	&60.90	&\textbf{67.31}	&60.90 &61.36	&44.66 &58.24	&59.24	&58.76	&\cellcolor[gray]{0.9}65.60 &\cellcolor[gray]{0.9}\textcolor{red}{-1.71}\\ 
        \textbf{BloodMNIST}	&12.66	&40.33	&33.75	&39.84 &42.78	&21.78 &38.17	&27.56 	&22.98	&\cellcolor[gray]{0.9}\textbf{70.15} &\cellcolor[gray]{0.9}\textcolor{darkgreen}{+27.37} \\ 
        \textbf{TissueMNIST} &15.70	&22.68	&17.45	&20.81 &18.35	&18.76 &20.43	&20.87	&14.81	&\cellcolor[gray]{0.9}\textbf{27.34} &\cellcolor[gray]{0.9}\textcolor{darkgreen}{+4.66}\\ 
        \textbf{OrganAMNIST}	&24.99	&39.97	&46.39	&45.72 &39.46	&33.93 &47.27	&43.23	&36.60	&\cellcolor[gray]{0.9}\textbf{54.75} &\cellcolor[gray]{0.9}\textcolor{darkgreen}{+7.48}\\ 
        \textbf{OrganCMNIST}	&22.37	&40.91	&36.98	&40.22 &37.45	&28.52	&38.42	&39.02 &30.63	&\cellcolor[gray]{0.9}\textbf{49.22} &\cellcolor[gray]{0.9}\textcolor{darkgreen}{+8.31}\\ 
        \textbf{OrganSMNIST}	&24.47	&38.46	&33.93	&37.62 &34.68	&27.92 &32.18	&30.65	&28.92	&\cellcolor[gray]{0.9}\textbf{45.41} &\cellcolor[gray]{0.9}\textcolor{darkgreen}{+6.95}\\ 
       \midrule
       \textbf{Average} &31.32	&46.59	&48.15	&46.69 &47.38	&36.87 &47.66	&46.26	&41.37	&\cellcolor[gray]{0.9}\textbf{56.51} &\cellcolor[gray]{0.9}\textcolor{darkgreen}{+8.36}\\
    
    \bottomrule
        
    \end{tabular}}
    \vspace{-0.2cm}
    \label{tab:fewshot}
\end{table*}

\begin{table*}[!ht]
    \centering
    \caption{\textbf{Comparison of methods on the Base-to-New generalization task of MedMNIST+ benchmark.}}
    \vspace{-0.25cm}
    \scalebox{0.58}{
    \begin{tabular}{lc|ccccccccc|cc}
    \toprule
        & &\multicolumn{1}{c}{BioMedCLIP} &\multicolumn{1}{c}{CoOp} &\multicolumn{1}{c}{CoCoOp} &\multicolumn{1}{c}{KgCoOp} &\multicolumn{1}{c}{MaPLe} &\multicolumn{1}{c}{ProGrad} &\multicolumn{1}{c}{PromptSRC} &\multicolumn{1}{c}{TCP} &\multicolumn{1}{c}{BioMedCoOp}& BioVLM &$\Delta$\\
        
        \multirow{-2}{*}{\textbf{Dataset}}  & \multirow{-2}{*}{\textbf{Sets}} &NEJM AI'25 &IJCV'22 &CVPR'22 &CVPR'23 &CVPR'23 &ICCV'23 &ICCV'23 &CVPR'24 &CVPR'25 & (Ours) & (in \%) \\ 
        \midrule

        &Base	&36.45 &42.32 &41.62 &42.65 &43.42	&35.90 &43.49	&43.03 &40.01 &\cellcolor[gray]{0.9}\textbf{57.84} &\cellcolor[gray]{0.9}\textcolor{darkgreen}{+14.35} \\ 
        
        &New	&43.49 &44.97 &42.20 &44.86 &44.67		 &44.54 &45.35	&44.39 &45.40 &\cellcolor[gray]{0.9}\textbf{46.86} &\cellcolor[gray]{0.9}\textcolor{darkgreen}{+1.46} \\ 
        
         \multirow{-3}{*}{\parbox{2.2cm}{\centering\textbf{Average on}\\\textbf{9 datasets}}} &H &39.66 &43.60 &41.91 &43.73 &44.04 &39.76 &44.40	&43.70 &42.54 &\cellcolor[gray]{0.9}\textbf{51.77} &\cellcolor[gray]{0.9}\textcolor{darkgreen}{+7.37}\\
         \midrule
         \midrule

         &Base &56.62	&86.13	&70.42	&85.13 &78.91	&53.68 &80.31	&75.21	&74.59	&\cellcolor[gray]{0.9}\textbf{92.73} &\cellcolor[gray]{0.9}\textcolor{darkgreen}{+6.60}\\
         
         &New	&45.83	&59.08	&47.35	&60.03 &55.23	&57.40 &52.45	&57.12	&56.56	&\cellcolor[gray]{0.9}\textbf{63.39} &\cellcolor[gray]{0.9}\textcolor{darkgreen}{+3.36}\\
         
         \multirow{-3}{*}{\textbf{PathMNIST}} &H &50.66	&70.09	&56.63	&70.41	&64.98 &55.48 &63.46 &64.93	&64.33	&\cellcolor[gray]{0.9}\textbf{75.30} &\cellcolor[gray]{0.9}\textcolor{darkgreen}{+4.89}\\
         \midrule

         &Base	&26.70	&34.06	&32.85	&35.60 &30.58	&36.16 &34.15	&29.60	&28.23	&\cellcolor[gray]{0.9}\textbf{52.83} &\cellcolor[gray]{0.9}\textcolor{darkgreen}{+16.67} \\
         
         &New	&62.08	&83.76	&82.93	&83.60 &82.59	&82.63 &83.12	&82.32	&81.17	&\cellcolor[gray]{0.9}\textbf{83.87} &\cellcolor[gray]{0.9}\textcolor{darkgreen}{+0.11}\\
         
         \multirow{-3}{*}{\textbf{DermaMNIST}} &H &37.34	&48.43	&47.06	&49.93 &44.63	&50.31 &48.41	&43.54	&41.89	&\cellcolor[gray]{0.9}\textbf{64.83} &\cellcolor[gray]{0.9}\textcolor{darkgreen}{+14.52} \\
         \midrule

         &Base	&57.80	&70.47	&75.53	&70.27	&75.82 &65.27 &\textbf{76.07}	&75.28	&69.40	&\cellcolor[gray]{0.9}75.33 &\cellcolor[gray]{0.9}\textcolor{red}{-0.74}\\
         
         &New	&\textbf{52.80}	&50.93	&48.20	&50.67 &49.52	&51.00 &47.10	&48.25	&48.60	&\cellcolor[gray]{0.9}46.40 &\cellcolor[gray]{0.9}\textcolor{red}{-6.40}\\
         
         \multirow{-3}{*}{\textbf{OCTMNIST}} &H &55.19	&59.13	&58.85	&58.88	&\textbf{59.91} &57.26 &58.18	&58.81	&57.17	&\cellcolor[gray]{0.9}57.43 &\cellcolor[gray]{0.9}\textcolor{red}{-2.48}\\
         \midrule

         &Base	&45.83	&36.00	&44.87	&38.46	&44.29 &31.52 &42.50	&41.98	&40.81	&\cellcolor[gray]{0.9}\textbf{49.14} &\cellcolor[gray]{0.9}\textcolor{darkgreen}{+3.31}\\
         
         &New	&39.77	&55.15	&59.47	&53.26	&60.74 &\textbf{62.54} &62.06	&62.87	&57.74	&\cellcolor[gray]{0.9}61.74 &\cellcolor[gray]{0.9}\textcolor{red}{-0.80}\\
         
         \multirow{-3}{*}{\textbf{RetinaMNIST}} &H &42.59 &43.57	&51.15	&44.67	&51.23 &41.91	&50.45	&50.34 &47.82	&\cellcolor[gray]{0.9}\textbf{54.73} &\cellcolor[gray]{0.9}\textcolor{darkgreen}{+3.50}\\
         \midrule

         &Base	&36.63	&33.16	&30.96	&34.15	&34.83 &35.11 &35.51	&36.02	&32.35	&\cellcolor[gray]{0.9}\textbf{68.94} &\cellcolor[gray]{0.9}\textcolor{darkgreen}{+32.31}\\
         
         &New	&23.81	&41.21	&27.22	&42.20	&42.48 &37.00 &46.23	&40.72	&50.07	&\cellcolor[gray]{0.9}\textbf{51.19} &\cellcolor[gray]{0.9}\textcolor{darkgreen}{+1.12}\\
         
         \multirow{-3}{*}{\textbf{BloodMNIST}} &H &28.86	&36.75	&28.97	&37.75	&38.28 &36.03 &40.17	&38.23	&39.31	&\cellcolor[gray]{0.9}\textbf{58.76} &\cellcolor[gray]{0.9}\textcolor{darkgreen}{+18.59}\\
         \midrule
         \midrule

         &Base	&6.53	&16.52	&12.27	&15.62	&20.48 &7.58 &17.43	&21.70	&9.79	&\cellcolor[gray]{0.9}\textbf{32.84} &\cellcolor[gray]{0.9}\textcolor{darkgreen}{+12.36}\\
         
         &New	&\textbf{36.66}	&26.22	&20.09	&25.86 &19.32	&22.12 &23.98	&18.75	&22.90	&\cellcolor[gray]{0.9}19.03 &\cellcolor[gray]{0.9}\textcolor{red}{-17.63}\\
         
         \multirow{-3}{*}{\textbf{TissueMNIST}} &H &11.09	&20.27	&15.23	&19.47	&19.88 &11.29 &20.19	&20.12	&13.71	&\cellcolor[gray]{0.9}\textbf{24.10} &\cellcolor[gray]{0.9}\textcolor{darkgreen}{+3.83}\\
         \midrule

         &Base	&29.53	&31.99	&33.38	&32.31 &35.28	&28.76 &32.37	&36.81	&33.71	&\cellcolor[gray]{0.9}\textbf{48.00} &\cellcolor[gray]{0.9}\textcolor{darkgreen}{+11.19}\\
         
         &New	&\textbf{48.40}	&32.65	&33.63	&33.15 &29.14	&33.23 &35.18	&27.05	&35.38	&\cellcolor[gray]{0.9}32.97 &\cellcolor[gray]{0.9}\textcolor{red}{-15.43}\\
         
         \multirow{-3}{*}{\textbf{OrganAMNIST}} &H &36.68	&32.31	&33.50	&32.72	&31.92 &30.84 &33.72	&31.18	&34.52	&\cellcolor[gray]{0.9}\textbf{39.09} &\cellcolor[gray]{0.9}\textcolor{darkgreen}{+2.41}\\
         \midrule

         &Base	&32.67	&36.81	&37.67	&37.00	&35.46 &32.59 &36.02	&34.19	&36.01	&\cellcolor[gray]{0.9}\textbf{52.17} &\cellcolor[gray]{0.9}\textcolor{darkgreen}{+14.50}\\
         
         &New	&\textbf{40.04}	&31.65	&32.07	&29.92	&34.67 &27.84 &32.54	&35.40	&27.93	&\cellcolor[gray]{0.9}33.17 &\cellcolor[gray]{0.9}\textcolor{red}{-6.87}\\
         
         \multirow{-3}{*}{\textbf{OrganCMNIST}} &H &35.98	&34.04	&34.65	&33.08	&35.06 &30.03 &34.19	&34.78	&31.46	&\cellcolor[gray]{0.9}\textbf{40.55} &\cellcolor[gray]{0.9}\textcolor{darkgreen}{+4.57}\\
         \midrule

         &Base	&35.73	&35.71	&36.67	&35.35	&35.13 &32.44 &37.08	&36.49	&35.24	&\cellcolor[gray]{0.9}\textbf{48.54} &\cellcolor[gray]{0.9}\textcolor{darkgreen}{+11.46}\\
         
         &New	&\textbf{42.01}	&24.03	&28.83	&25.06 &28.38	&27.09	&25.45	&27.03 &28.29	&\cellcolor[gray]{0.9}29.98 &\cellcolor[gray]{0.9}\textcolor{red}{-12.03}\\
         
         \multirow{-3}{*}{\textbf{OrganSMNIST}} &H &\textbf{38.62}	&28.73	&32.28	&29.33 &31.40	&29.52	&30.18	&31.06 &31.39	&\cellcolor[gray]{0.9}37.06 &\cellcolor[gray]{0.9}\textcolor{red}{-1.56}\\
    
    \bottomrule
        
    \end{tabular}}
    \label{tab:b2nmedmnist}
\end{table*}
\section{Experimental Results}
\label{sec:experiments}

\vspace{-0.2cm}
\noindent \textbf{Dataset.} We evaluate our method on 11 diverse 2D medical imaging datasets from the MedMNIST+ benchmark~\cite{medmnist}, covering nine imaging modalities and a wide range of anatomical structures and clinical protocols. The multi-label ChestMNIST dataset is excluded from the 2D setting. For all experiments, we use the standard 224×224 resolution images. A detailed description of each dataset is available in the \texttt{Appendix}.

\noindent \textbf{Implementation Details.} We conduct all experiments on a single NVIDIA A6000 GPU, using ViT-B/16 from pretrained BioMedCLIP~\cite{BioMedCLIP} as the backbone and GPT-4o~\cite{gpt4} for attribute generation. We report results averaged over three independent runs and train all models for 50 epochs across generalization settings. We set $N=10$ prompts per category, each of length $M=4$, with $\lambda_1=\lambda_3=1$ and $\lambda_2=0.5$. We train using SGD with a learning rate of $2\times10^{-3}$, batch size 32, and $\mathcal{K}=16$ shots, together with a cosine scheduler and a 1-epoch warm-up at $1\times10^{-5}$. We apply strong augmentations including random horizontal flip, color jitter, and Gaussian blur, while weak augmentations use horizontal flip and $10^\circ$ rotation. 

\subsection{Comparison with the state-of-the-art methods}
\vspace{-0.2cm}
\noindent \textbf{Few-shot learning:} We evaluate BioVLM on the few-shot learning (FSL) task, considering only textual prompt learning methods as baselines for fair comparison~\cite{biomedcoop}. As shown in Table~\ref{tab:fewshot}, BioVLM consistently outperforms prior methods, achieving at least an 8.36\% gain over the second-best baseline. By leveraging entropy-based prompt selection and LLM-driven attribute alignment, our proposed BioVLM effectively addresses challenges such as class imbalance and low inter-class variance. Compared to deeper prompting approaches e.g., MaPLe, PromptSRC, TCP, it exhibits stronger generalization with lower complexity. Notably, BioVLM outperforms CoOp by 29.82\% on BloodMNIST and TissueMNIST. While gains are smaller on RetinaMNIST and BreastMNIST due to dataset bias, BioVLM shows strong improvements in Base-to-New generalization.

\noindent \textbf{Base-to-New (B2N) Generalization:} 
We evaluate the model’s ability to recognize novel categories within the same imaging modality, a stringent test of semantic generalization (Table~\ref{tab:b2nmedmnist}). BioVLM achieves the highest harmonic mean (HM), outperforming prior baselines by an average of +8.04\% across PathMNIST, DermaMNIST, and BloodMNIST. These gains demonstrate the effectiveness of entropy-guided prompt selection and attribute-aligned representations. PneumoniaMNIST and BreastMNIST are excluded, as each contains only one class in either the base or novel split.

The performance remains challenging on TissueMNIST and OrganMNIST due to pronounced intra-modality variation, where novel classes differ structurally and functionally from base classes (e.g., ``bladder'' vs. ``lung''). This can limit prompt specialization and reduce performance relative to zero-shot BioMedCLIP. Nonetheless, BioVLM narrows this gap, confirming the utility of its LLM-grounded multi-prompt strategy. These results suggest future directions such as integrating domain-invariant regularization or hierarchical priors to handle anatomically diverse tasks.
\begin{table*}[htbp]
    \centering
    \caption{\textbf{Comparison of methods on the out-of-distribution generalization task of MedMNIST+ benchmark.}}
    \vspace{-0.25cm}
    \scalebox{0.6}{
    \begin{tabular}{l|cccccccc|cc}
    \toprule
         &\multicolumn{1}{c}{CoOp} &\multicolumn{1}{c}{CoCoOp} &\multicolumn{1}{c}{KgCoOp} &\multicolumn{1}{c}{MaPLe} &\multicolumn{1}{c}{ProGrad} &\multicolumn{1}{c}{PromptSRC} &\multicolumn{1}{c}{TCP} &\multicolumn{1}{c}{BioMedCoOp}& BioVLM &$\Delta$ \\
        
        \multirow{-2}{*}{\textbf{Source}} &IJCV'22 &CVPR'22 &CVPR'23 &CVPR'23 &ICCV'23 &ICCV'23 &CVPR'24 &CVPR'25 & (Ours) & (in \%) \\ 
        \midrule

        \textbf{PathMNIST}	&31.94 &29.68 &32.43 &32.51 &32.08 &32.11 &32.18 &\textbf{32.72} &\cellcolor[gray]{0.9}31.55 & \cellcolor[gray]{0.9}\textcolor{red}{-1.17}\\ 
        \textbf{DermaMNIST}	&22.73 &26.67 &23.29 &22.79 &22.86 &24.42 &23.17 &29.38 &\cellcolor[gray]{0.9}\textbf{29.41} &\cellcolor[gray]{0.9}\textcolor{darkgreen}{+0.03} \\ 
        \textbf{OCTMNIST}	&31.56 &30.53 &31.50 &31.95 &31.03 &32.43 &32.47 &31.45 &\cellcolor[gray]{0.9}\textbf{32.89} &\cellcolor[gray]{0.9}\textcolor{darkgreen}{+0.42}\\ 
        \textbf{PneumoniaMNIST}	&21.18 &26.05 &22.12 &25.39 &23.34 &25.01 &26.15 &26.89 &\cellcolor[gray]{0.9}\textbf{28.19} &\cellcolor[gray]{0.9}\textcolor{darkgreen}{+1.30} \\ 
        \textbf{RetinaMNIST}	&\textbf{34.68} &30.06 &33.18 &31.68 &33.65 &32.24 &31.99 &32.92 &\cellcolor[gray]{0.9}31.37 &\cellcolor[gray]{0.9}\textcolor{red}{-3.31} \\ 
        
        \textbf{BreastMNIST} &29.27 &25.54 &29.78 &30.16 &30.14 &\textbf{31.02} &30.75 &30.35 &\cellcolor[gray]{0.9}28.93 &\cellcolor[gray]{0.9}\textcolor{red}{-2.09} \\ 
        
        \textbf{BloodMNIST} &29.57 &28.45 &28.50 &31.32 &28.84 &32.60 &31.80 &28.21 &\cellcolor[gray]{0.9}\textbf{30.38} &\cellcolor[gray]{0.9}\textcolor{darkgreen}{+0.81} \\ 
        
        \textbf{TissueMNIST} &31.06 &30.73 &34.88 &32.96 &33.60 &34.14 &32.58 &34.91 &\cellcolor[gray]{0.9}\textbf{36.28} &\cellcolor[gray]{0.9}\textcolor{darkgreen}{+1.40} \\ 
        \textbf{OrganAMNIST}	&33.83 &\textbf{34.70} &33.02 &32.44 &34.68 &33.30 &33.80 &33.51 &\cellcolor[gray]{0.9}34.34 &\cellcolor[gray]{0.9}\textcolor{red}{-0.36} \\ 
        \textbf{OrganCMNIST}	& 34.20 &35.47 &33.55 &34.09 &33.83 &34.23 &34.55 &32.62 &\cellcolor[gray]{0.9}\textbf{36.57} &\cellcolor[gray]{0.9}\textcolor{darkgreen}{+1.10} \\ 
        \textbf{OrganSMNIST}	&34.78 &32.38 &35.04 &33.33 &34.32 &33.24 &32.68 &33.74 &\cellcolor[gray]{0.9}\textbf{35.83} &\cellcolor[gray]{0.9}\textcolor{darkgreen}{+0.79} \\ 
       \midrule
       
       \textbf{Average} &30.44 &30.02 &30.66 &30.78 &30.76 &31.34 &31.10 &31.52 &\cellcolor[gray]{0.9}\textbf{32.34} &\cellcolor[gray]{0.9}\textcolor{darkgreen}{+0.82} \\
    
    \bottomrule
        
    \end{tabular}}
    \vspace{-0.2cm}
    \label{tab:ood}
\end{table*}

\noindent \textbf{Out-of-distribution Generalization:}
In the OOD generalization setting, we train models on a single source dataset and evaluate them on all remaining datasets~\cite{stylip}. Table~\ref{tab:ood} reports the average accuracy across unseen target domains. BioVLM achieves the best overall performance, outperforming all prompt-learning baselines by +0.82\% over the second-best method. Although margins are smaller than in other settings, BioVLM shows consistent generalization across diverse unseen medical imaging tasks, indicating that its entropy-based prompt selection and LLM-derived attributes yield domain-robust representations. Compared to CoOp and CoCoOp, which often overfit to domain-specific artifacts, BioVLM maintains strong performance on structurally diverse datasets such as RetinaMNIST, TissueMNIST, and OrganMNIST. While minor drops are observed on PathMNIST and BreastMNIST due to large domain shifts, the results confirm BioVLM’s strong OOD transfer ability. Detailed per-source results are provided in the \texttt{Appendix}.

\subsection{Ablation Experiments}
\vspace{-0.2cm}

\noindent\textbf{Effect of different loss components:} Table~\ref{tab:loss} presents an ablation study on Base-to-New generalization across all datasets, analyzing the impact of different loss components in Eq.~\ref{eq:total_loss}. Using only cross-entropy loss ($\mathcal{L}_{\text{CE}}$) yields suboptimal performance (HM 45.01\%), indicating that standard classification objectives are insufficient for robust biomedical generalization. Adding $\mathcal{L}_{\text{LER}}$ significantly improves performance, with gains of 5.10\% and 2.38\% on base and novel classes, respectively, highlighting its role in filtering uncertain prompts and promoting confident predictions. Incorporating $\mathcal{L}_{\text{CMD}}$ further boosts HM, demonstrating effective semantic transfer from LLM-derived embeddings to learned prompts. The full loss formulation defined in Eq. \ref{eq:total_loss} achieves the best performance, including an additional 0.74\% improvement on novel classes. Similar trends are observed in few-shot learning, where combining all loss terms consistently improves accuracy, with performance increasing as the number of shots grows.

\begin{table}[!ht]
    \centering
    \caption{\textbf{Effect of the loss functions on B2N generalization and FSL tasks. $\mathcal{K}$ is
    the number of shots. }}
    \vspace{-0.3cm}
    \scalebox{0.53}{
    \begin{tabular}{lccc|ccc|cccc}
    \toprule
        $\mathcal{L}_{CE}$ & $\mathcal{L}_{LER}$ & $\mathcal{L}_{CMD}$ & $\mathcal{L}_{ASA}$ & Base & Novel & HM &$\mathcal{K}$=1	&$\mathcal{K}$=4	&$\mathcal{K}$=8	&$\mathcal{K}$=16 \\
        \midrule
        \centering$\checkmark$	& \centering$\times$ &\centering$\times$ &\centering$\times$ &46.82	&43.34	&45.01 &35.07	&41.27	&47.82	&53.26 \\ 
        
        \centering$\checkmark$	& \centering$\checkmark$ &\centering$\times$ &\centering$\times$ &51.92	&45.72	&48.62 &37.21	&43.11	&48.72	&54.67 \\

        \centering$\checkmark$	& \centering$\times$ &\centering$\checkmark$ &\centering$\times$ &52.74	&44.65	&48.36 &36.24	&42.89	&48.14	&53.98 \\

        \centering$\checkmark$	& \centering$\times$ &\centering$\times$ &\centering$\checkmark$ &50.56	&44.10	&47.11 &34.78	&41.66	&46.22	&53.78\\

        \centering$\checkmark$	& \centering$\checkmark$ &\centering$\checkmark$ &\centering$\times$ &54.48	&45.37	&49.51 &39.61	&47.59	&51.35	&56.05 \\

        \centering$\checkmark$	& \centering$\times$ &\centering$\checkmark$ &\centering$\checkmark$ &55.20	&45.82	&50.07 &38.56	&45.72	&51.19	&55.34 \\

        \centering$\checkmark$	& \centering$\checkmark$ &\centering$\times$ &\centering$\checkmark$ &55.93	&46.12	&50.55 &39.87	&46.45	&50.78	&55.92 \\

       \midrule
       \centering$\checkmark$	& \centering$\checkmark$ &\centering$\checkmark$ &\centering$\checkmark$ &\textbf{57.84} &\textbf{46.86} &\textbf{51.77} &\textbf{40.23}	&\textbf{47.94}	&\textbf{51.70}	&\textbf{56.51} \\
    \bottomrule       
    \end{tabular}}
    \label{tab:loss}
\end{table}

\begin{figure}
    \centering
    \includegraphics[width=\columnwidth]{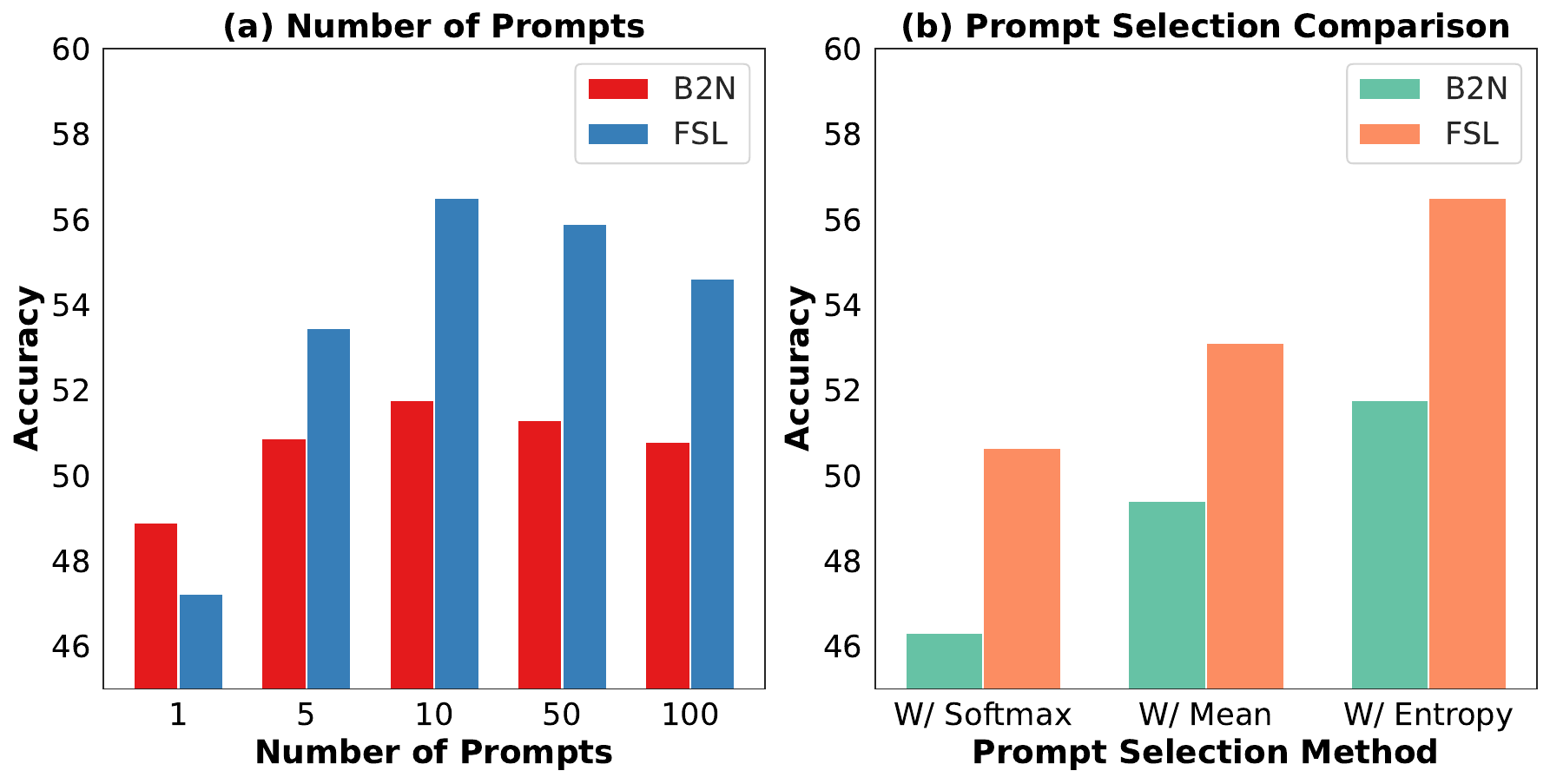}
    \vspace{-0.6cm}
    \caption{\textbf{Model ablation: Varying (a) number of prompts, (b) prompt selection methods (W/ = with).}}
    \label{fig:plots}
    \vspace{-0.4cm}
\end{figure}

\noindent\textbf{Ablation with number of learnable prompts:} To study the effect of prompt diversity, we evaluate performance with varying numbers of learnable prompts per class (Figure~\ref{fig:plots}(a)), reporting averages across all datasets. Increasing prompts from 1 to 10 steadily improves accuracy, from 48.9\% to 51.77\% on Base-to-New generalization and from 47.22\% to 56.51\% on FSL. This demonstrates that greater prompt diversity enhances generalization by providing multiple contextual views per class. However, gains saturate beyond 10 prompts, with performance slightly degrading at 50 and 100 prompts for example, B2N drops to 50.79\% at 100, likely due to redundant or noisy prompt variants.

\noindent\textbf{Performance against prompt selection methods:} BioVLM enhances generalization by selecting the most informative prompts from a prompt pool. We compare entropy-based selection with softmax- and mean-based strategies (Figure~\ref{fig:plots}(b)). The entropy-based approach consistently outperforms softmax by 5.45\% (Base-to-New) and 5.86\% (FSL), and mean-based selection by 2.33\% and 3.36\%, respectively, averaged across all datasets. Additional ablations on $\lambda$ sensitivity and the impact of different LLMs are provided in the \texttt{Appendix}.

\noindent\textbf{Computational complexity:}Table \ref{tab:params_m} demonstrates that BioVLM attains superior performance in base-to-new (B2N) generalization and fewshot learning (FSL) settings, while utilizing substantially fewer trainable parameters compared to existing methods such as MaPLe, PromptSRC, and TCP.\vspace{-0.1cm}
\begin{table}[ht]
\centering
\caption{\textbf{Ablation with the trainable parameters}.}
\vspace{-0.3cm}
\scalebox{0.7}{
\begin{tabular}{lccc}
\hline

\textbf{Method} & \textbf{Params. (in Millions)} & \textbf{B2N} & \textbf{FSL} \\
\hline
CoOp        & 0.0031 & 43.60 & 46.59 \\
CoCoOp      & 0.0448 & 41.91 & 48.15 \\
KgCoOp      & 0.0031 & 43.73 & 46.69 \\
MaPLe       & 5.3460 & 44.04 & 47.38 \\
ProGrad     & 0.0031 & 39.76 & 36.87 \\
PromptSRC   & 0.0690 & 44.40 & 47.66 \\
TCP         & 0.4965 & 43.70 & 46.26 \\
BioMedCoOp  & 0.0031 & 42.54 & 41.37 \\
BioVLM (Ours)    & 0.0307 & \textbf{51.77} & \textbf{56.51} \\
\hline
\end{tabular}
}\vspace{-0.4cm}
\label{tab:params_m}
\end{table}

\section{Conclusions}
We introduce BioVLM, a prompt-based framework that significantly enhances the generalization of biomedical vision-language models. BioVLM, a novel framework, works by dynamically selecting the most informative prompts from a diverse pool, synergistically combining attribute alignment, entropy regularization, and knowledge distillation, guides the model to learn confident and semantically grounded representations. Building on the results, 
future work can develop methods to automatically refine or reduce the dependency on LLM-generated attributes, and extending the framework to 3D volumetric data.

\section{Limitations}
BioVLM relies on LLM-generated attributes to guide prompt learning, which introduces additional computational overhead during the attribute extraction stage and makes the approach dependent on the quality of the generated descriptions, although this process is performed offline. In addition, our experiments focus on 2D medical imaging datasets, and extending the framework to volumetric modalities like  CT or MRI may require additional architectural and computational considerations. Finally, the use of multiple learnable prompts per class increases training complexity, which may affect efficiency in large-scale or highly diverse biomedical settings.

\bibliography{custom}

\appendix

\setcounter{figure}{0}
\renewcommand{\thefigure}{A\arabic{figure}}

\setcounter{table}{0}
\renewcommand{\thetable}{A\arabic{table}}
\clearpage
\section{Appendix}
\label{sec:appendix}

\begin{itemize}
\item [1.] \textbf{Overview of the MedMNIST+ datasets}: In Table~\ref{tab:data}, we provide a detailed overview of the datasets included in MedMNIST+, highlighting key attributes such as imaging modality, number of classes, total samples, and the standardized training, validation, and testing splits. MedMNIST+ comprises 11 publicly available 2D medical image classification datasets, namely PathMNIST, ChestMNIST, DermaMNIST, OCTMNIST, PneumoniaMNIST, RetinaMNIST, BreastMNIST, OrganAMNIST, OrganCMNIST, OrganSMNIST, and BloodMNIST. These datasets cover a diverse set of imaging modalities, including pathology slides, chest and hand X-rays, abdominal CT scans, dermoscopy, retinal fundus, and OCT imaging and support both binary and multi-class classification tasks. Their heterogeneity in anatomical focus, class granularity, and domain distribution makes MedMNIST+ a comprehensive benchmark for evaluating the robustness and generalization ability of vision-language models across different tasks and domains in medical imaging.

\item [2.] \textbf{Overview of the 11 biomedical datasets}: We also conduct experiments on 11 biomedical datasets spanning diverse imaging modalities, following the same train-validation-test splits as in \cite{biomedcoop}. These datasets include CTKidney \cite{ctkidney} for computed tomography; DermaMNIST \cite{derma1, derma2} for dermatoscopy; Kvasir \cite{kvasir} for endoscopy; RETINA \cite{retina1, retina2} for fundus photography; LC25000 \cite{lc25} and CHMNIST \cite{chmnist} for histopathology; BTMRI \cite{btmri} for magnetic resonance imaging; OCTMNIST \cite{octmnist} for optical coherence tomography; BUSI \cite{busi} for ultrasound; and COVID-QU-Ex \cite{covid19} and KneeXray \cite{kneexray} for X-ray imaging.

\begin{figure}[ht!]
    \centering
    \includegraphics[width=\columnwidth]{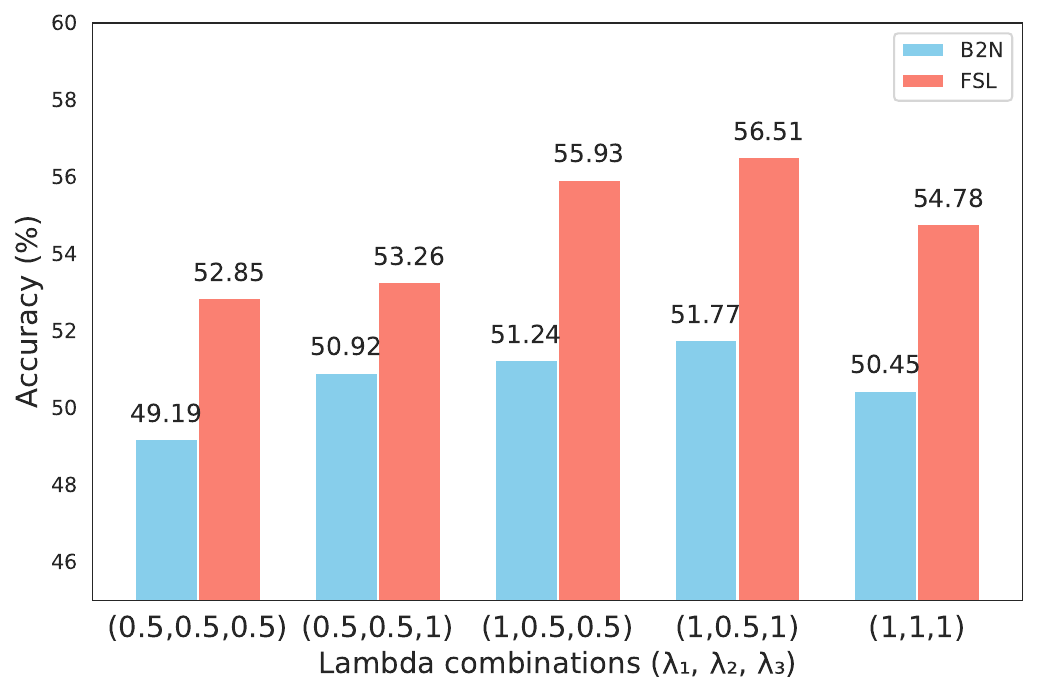}
    \caption{Ablation for the hyperparameter $\lambda_1$, $\lambda_2$ and $\lambda_3$ for the Base-to-New Generalization and Few-shot learning tasks.}
    \label{fig:lambda}
\end{figure}

\item [3.] \textbf{Few-shot learning and base-to-new generalization on the biomedical benchmark datasets:} Tables \ref{tab:fewshot_biomed} and \ref{tab:b2nbiomed} present the comparison of BioVLM with state-of-the-art baseline methods across the biomedical datasets used by BioMedCoOp \cite{biomedcoop} under few-shot learning and base-to-new generalization settings respectively. We consider CLIP-based adapter baselines as CLIP-Adapter \cite{clip-adapter}, and Tip-Adapter \cite{tip-adapter}, linear probing baselines as Standard LP \cite{clip}, LP++ \cite{lp++}, and prompt learning baselines same as Table \ref{tab:fewshot}. The results clearly demonstrate that BioVLM consistently outperforms competing approaches, highlighting its robustness and superior generalization capability in challenging biomedical scenarios.

\item [4.] \textbf{Model Calibration Performance:} Beyond classification accuracy, we analyze model calibration on the few-shot learning task across 11 biomedical datasets. using Expected Calibration Error (ECE) \cite{ece}. As shown in Table~\ref{tab:calibration}, our BioVLM demonstrates consistently stronger calibration than the compared methods across diverse biomedical modalities. In particular, BioVLM substantially improves over the zero-shot BiomedCLIP baseline, indicating that the proposed adaptation strategy leads to more reliable confidence estimates. Compared with existing prompt-learning approaches, BioVLM also shows more stable calibration behavior across datasets, whereas several baselines exhibit modality-dependent fluctuations. Although MaPLe remains competitive on a few datasets, BioVLM achieves the best overall calibration trend, suggesting that biomedical-specific visual-language modeling better captures uncertainty under few-shot supervision.

\item [5.] \textbf{$\lambda$-Hyperparameter Sensitivity:} We ablate our proposed BioVLM for the best combination of hyperparameters ($\lambda_1$, $\lambda_2$ and $\lambda_3$) defined in Eq.\ref{eq:total_loss} and showcase the results in Figure \ref{fig:lambda} for the Base-to-New Generalization and Few-shot learning tasks.

\item [6.] \textbf{Effect of different LLMs:} In Table~\ref{tab:llm}, we present the impact of different LLMs including Llama-3.2-3B \cite{llama3}, Qwen2.5-14B \cite{qwen2}, Phi-4 \cite{phi4} and GPT-4o \cite{gpt4o}, on the Base-to-New Generalization and Few-shot learning tasks. The results show that variations in the choice of LLM have minimal effect on our proposed method, with BioVLM consistently outperforming all baselines across both tasks.

\item [7.] \textbf{Ablation with additional prompt-selection methods:} Table~\ref{tab:prompt_select} compares different strategies for selecting or aggregating prompts from the prompt bank. In our method, each class is represented by multiple LLM-guided learnable prompts, and each prompt produces image-text similarity scores through the frozen encoders. \textit{Softmax} directly uses probability scores but may emphasize noisy high-confidence prompts, while \textit{Mean} averages all prompt predictions and can dilute discriminative cues. \textit{Average Logits} combines raw similarity logits before normalization, but does not explicitly filter uncertain prompts. \textit{Argmax} selects the single most confident prompt based on prediction probabilities, whereas \textit{Top-2} and \textit{Top-5} average predictions from the most confident prompts using the maximum class probability as the confidence score. However, these fixed selection strategies may not adapt well across images or modalities. In contrast, \textit{entropy} selection chooses low-uncertainty prompts, which are more confident and discriminative for each input. This better aligns with BioVLM’s prompt-routing design and achieves the strongest performance in both base-to-new generalization and few-shot learning.

\item [8.] \textbf{Comparison with PEFT methods.}
We compare BioVLM with representative PEFT methods on the few-shot learning task of the MedMNIST+ benchmark. As shown in Table~\ref{tab:peft}, \textbf{BioVLM} consistently outperforms all PEFT baselines, demonstrating stronger adaptation under limited supervision. Compared with LoRA~\cite{lora}, AdaLoRA~\cite{adalora}, LayerNorm~\cite{layernorm}, and BitFit~\cite{bitfit}, BioVLM better leverages biomedical visual-language alignment rather than only adapting model parameters. It also surpasses adapter-based methods such as CLIP-Adapter~\cite{clip-adapter} and Tip-Adapter~\cite{tip-adapter}, which show less consistent performance across modalities, highlighting BioVLM as a more effective PEFT strategy for few-shot biomedical image recognition.

\item [9.] \textbf{Qualitative results:} Figure \ref{fig:tsne} presents the t-SNE visualization of the logits from BioMedCoOp and BioVLM on the PathMNIST dataset in the few-shot setting. The plot clearly shows that BioVLM achieves better class separation compared to BioMedCoOp.

\item [10.] \textbf{Details results on Out-of-Domain generalization task:} In Tables \ref{tab:fewshot_1} - \ref{tab:ood_no_organs}, we showcase the performance of Out-of-Domain (OOD) generalization task on 11 datasets, where our proposed BioVLM outperforms the state-of-the-art prompt learning methods by significant margin.
\end{itemize}

\begin{table}[!ht]
    \centering
    \caption{\textbf{Ablation of prompt selection methods in BioVLM on Base-to-New Generalization and Few-shot learning settings.}}
    \scalebox{0.88}{
    \begin{tabular}{l|cc}
    \toprule
        Method &B2N & FSL \\
        \midrule

        Softmax	&46.32	&50.65 \\
Mean	&49.62	&53.06 \\
Avg. Logits	&48.78	&52.57 \\
Argmax	&50.24	&54.19 \\
Top-2	&49.70	&53.56 \\
Top-5	&49.52	&53.40 \\
\midrule
\textbf{Entropy (ours)}	&51.77	&56.51 \\
    
    \bottomrule
    \end{tabular}}
    \label{tab:prompt_select}
\end{table}

\begin{table*}[!ht]
    \centering
    \caption{\textbf{Effect of different LLMs on base-to-new generalization and few-shot learning tasks.} We also showcase a snippet of attributes of the class `lymphocyte', belongs to the BloodMNIST dataset.}
    \vspace{-0.1cm}
    \scalebox{0.88}{
    \begin{tabular}{l|c|cc}
    \toprule
        Methods & Attribute Snippet & B2N & FSL \\
    \midrule
        Llama-3.2-3B &
        ``small round cell with the nucleus making up most of the cell volume''
        & 49.92 & 54.10 \\
        Qwen2.5-14B &
        ``small, rounded cell characterized by a high nuclear-to-cytoplasmic proportion''
        & 50.56 & 55.32 \\
        Phi-4 &
        ``small round cell''
        & 49.34 & 54.88 \\
        GPT-4o &``small, round cell with a high nuclear-to-cytoplasmic ratio'' & \textbf{51.77} & \textbf{56.51} \\
    \bottomrule
    \end{tabular}}
    \label{tab:llm}
\end{table*}

\begin{figure*}[h!]
    \centering
    \includegraphics[width=\textwidth]{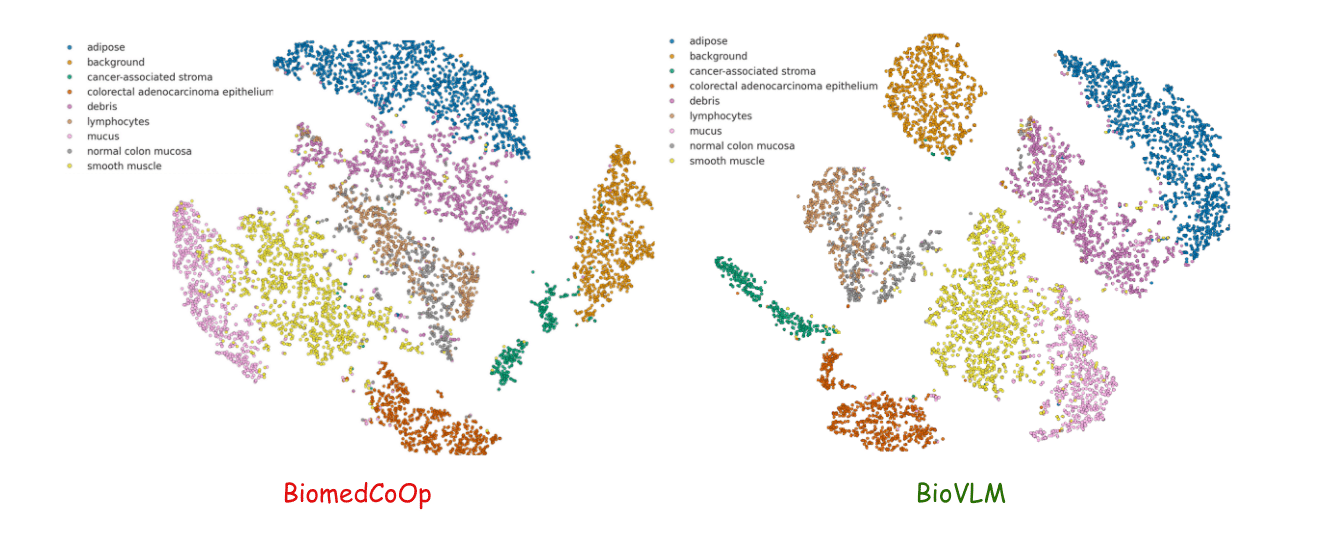}
    \caption{{t-SNE visualization of BioMedCoOp and BioVLM in few-shot evaluation on the PathMNIST dataset.}}
    \label{fig:tsne}
\end{figure*}

\begin{table*}[htbp!]

\caption{\textbf{Overview of the MedMNIST+ datasets, their modalities, classification tasks, number of samples, and dataset splits.}}
\centering
\scalebox{0.73}{
\begin{tabular}{l|l|l|l|l}
    \toprule
    \textbf{Dataset} & \textbf{Data Modality} & \textbf{Classes} & \textbf{\# Samples} & \textbf{\# Training / Validation / Test} \\
    \midrule
    PathMNIST & Colon Pathology &
    \begin{tabular}[l]{@{}l@{}}
    adipose, background, debris, \\lymphocytes, mucus,
    smooth muscle,\\ normal colon mucosa, \\
    cancer-associated stroma,\\ colorectal adenocarcinoma epithelium
    \end{tabular} &
    107,180 & 89,996 / 10,004 / 7,180 \\

    \midrule
    DermaMNIST & Dermatoscope &
    \begin{tabular}[l]{@{}l@{}}
    actinic keratoses and intraepithelial \\carcinoma, basal cell carcinoma, \\
    benign keratosis, dermatofibroma,\\ melanoma, 
    melanocytic nevi, \\vascular lesions
    \end{tabular} &
    10,015 & 7,007 / 1,003 / 2,005 \\

    \midrule
    OCTMNIST & Retinal OCT &
    \begin{tabular}[l]{@{}l@{}}
    choroidal neovascularization,\\ diabetic macular edema, 
    drusen, normal
    \end{tabular} &
    109,309 & 97,477 / 10,832 / 1,000 \\

    \midrule
    PneumoniaMNIST & Chest X-Ray & normal, pneumonia & 5,856 & 4,708 / 524 / 624 \\

    \midrule
    RetinaMNIST & Fundus Camera &
    \begin{tabular}[l]{@{}l@{}}
    normal retina, mild diabetic retinopathy, \\
    moderate diabetic retinopathy, \\
    severe diabetic retinopathy, \\proliferative diabetic retinopathy
    \end{tabular} &
    1,600 & 1,080 / 120 / 400 \\

    \midrule
    BreastMNIST & Breast Ultrasound & malignant, benign & 780 & 546 / 78 / 156 \\

    \midrule
    BloodMNIST & Blood Cell Microscope &
    \begin{tabular}[l]{@{}l@{}}
    basophil, eosinophil, \\erythroblast, immature granulocytes \\
    (myelocytes, metamyelocytes and \\promyelocytes), \\
    lymphocyte, monocyte, neutrophil, platelet
    \end{tabular} &
    17,092 & 11,959 / 1,712 / 3,421 \\

    \midrule
    TissueMNIST & Kidney Cortex Microscope &
    \begin{tabular}[l]{@{}l@{}}
    Collecting Duct, Connecting Tubule, \\Distal Convoluted Tubule, \\
    Glomerular endothelial cells,\\ Interstitial endothelial cells, \\
    Leukocytes, Podocytes,\\ Proximal Tubule Segments, \\
    Thick Ascending Limb
    \end{tabular} &
    236,386 & 165,466 / 23,640 / 47,280 \\

    \midrule
    OrganAMNIST & Abdominal CT &
    \begin{tabular}[l]{@{}l@{}}
    bladder, femur-left, femur-right,\\ heart, kidney-left, \\
    kidney-right, liver, lung-left,\\ lung-right,
    pancreas, spleen
    \end{tabular} &
    58,850 & 34,561 / 6,491 / 17,778 \\

    \midrule
    OrganCMNIST & Abdominal CT &
    \begin{tabular}[l]{@{}l@{}}
    bladder, femur-left, femur-right, heart,\\ kidney-left,
    kidney-right,\\ liver, lung-left, lung-right, \\
    pancreas, spleen
    \end{tabular} &
    23,583 & 12,975 / 2,392 / 8,216 \\

    \midrule
    OrganSMNIST & Abdominal CT &
    \begin{tabular}[l]{@{}l@{}}
    bladder, femur-left, femur-right, heart,\\ kidney-left,
    kidney-right, liver,\\ lung-left, lung-right, 
    pancreas, spleen
    \end{tabular} &
    25,211 & 13,932 / 2,452 / 8,827 \\
    \bottomrule
\end{tabular}
 }
\label{tab:data}
\end{table*}

\begin{table*}[!ht]
    \centering
    \caption{{\textbf{Comparison of methods on the few-shot learning task of the average of 11 biomedical datasets. }}}
    \vspace{-0.3cm}
    \scalebox{0.8}{
    \begin{tabular}{lccccc}
    \toprule
        \textbf{Method} &$K=1$ &$K=2$	&$K=4$	&$K=8$	&$K=16$ \\
        \midrule
        \rowcolor{cyan!10}\multicolumn{6}{c}{Zero-shot Methods} \\
        BioMedCLIP & & &42.05 & & \\
        BioMedCLIP + Ensemble & & &52.27 & & \\
        BioMedCLIP + Selective Ensemble & & &53.72 & & \\
       \midrule
       \rowcolor{cyan!10}\multicolumn{6}{c}{CLIP-based Adapter Methods} \\
       CLIP-Adapter &44.66 &43.91 &44.36 &45.42 &46.69 \\
       Tip-Adapter &49.19 &52.36 &57.33 &61.98 &67.15 \\
       Tip-Adapter-F &51.17 &52.74 &61.23 &65.91 &70.91 \\
       \midrule
       \rowcolor{cyan!10}\multicolumn{6}{c}{Linear Probing Methods} \\
       Standard LP &47.25 &54.21 &61.00 &65.85 &69.40 \\
       LP++ &47.24 &53.18 &59.02 &63.69 &68.35\\
       \midrule
       \rowcolor{cyan!10}\multicolumn{6}{c}{Prompt Learning Methods} \\
       CoOp &50.16 &54.18 &59.75 &65.84 &69.62 \\
       CoCoOp &48.49 &51.28 &54.69 &61.08 &65.09 \\
       KgCoOp &50.85 &53.18 &57.82 &62.08 &62.84 \\
       MaPLe &52.30	&56.17	&64.10	&69.29	&70.48 \\
       ProGrad &51.88 &54.71 &60.42 &65.61 &67.13 \\
       PromptSRC	&57.54	&59.73	&63.27	&68.36	&73.25 \\
       TCP	&55.68	&60.06	&64.35	&70.57	&71.80 \\
       BioMedCoOp &57.03 &59.13 &63.95 &68.32 &72.42 \\
       \midrule
       \cellcolor[gray]{0.9} BioVLM (Ours)	&\cellcolor[gray]{0.9}\textbf{59.91}	&\cellcolor[gray]{0.9}\textbf{62.34}	&\cellcolor[gray]{0.9}\textbf{66.50}	&\cellcolor[gray]{0.9}\textbf{72.86}	&\cellcolor[gray]{0.9}\textbf{77.62} \\
       \midrule
       \cellcolor[gray]{0.9} $\Delta$ (in \%)	&\cellcolor[gray]{0.9}\color{darkgreen}+2.37	&\cellcolor[gray]{0.9}\color{darkgreen}+2.28	&\cellcolor[gray]{0.9}\color{darkgreen}+2.15	&\cellcolor[gray]{0.9}\color{darkgreen}+2.29	&\cellcolor[gray]{0.9}\color{darkgreen}+4.37 \\
       
    \bottomrule       
    \end{tabular}}
    \vspace{-0.1cm}
    \label{tab:fewshot_biomed}
\end{table*}

\begin{table*}[!ht]
    \centering
    \caption{{\textbf{Comparison of methods on the Base-to-New generalization task of 10 biomedical datasets.}}}
    \vspace{-0.2cm}
    \scalebox{0.59}{
    \begin{tabular}{cc|ccccccccc|>{\columncolor[gray]{0.9}}c>{\columncolor[gray]{0.9}}c}
    \toprule
        & &\multicolumn{1}{c}{BioMedCLIP} &\multicolumn{1}{c}{CoOp} &\multicolumn{1}{c}{CoCoOp} &\multicolumn{1}{c}{KgCoOp} &\multicolumn{1}{c}{MaPLe} &\multicolumn{1}{c}{ProGrad} &\multicolumn{1}{c}{PromptSRC} &\multicolumn{1}{c}{TCP} &\multicolumn{1}{c}{BioMedCoOp} & BioVLM & $\Delta$\\
        \multirow{-2}{*}{\textbf{Dataset}}  & \multirow{-2}{*}{\textbf{Sets}} &NEJM AI'25 &IJCV'22 &CVPR'22 &CVPR'23 &CVPR'23 &ICCV'23 &ICCV'23 &CVPR'24 &CVPR'25 & (Ours) & (in \%)  \\ 
        \midrule

         &Base &47.84 &73.85 &72.26 &68.36 &73.05 &71.67 &75.28 &73.82 &76.26 &\textbf{77.19} & {\color{darkgreen}+0.93} \\
         &New  &65.42 &64.75 &67.03 &64.08 &71.46 &66.93 &71.33 &70.74 &73.92 &\textbf{75.15} & {\color{darkgreen}+1.23} \\
         \multirow{-2}{*}{\parbox{2.2cm}{\centering\textbf{Average on 10 Datasets}}}&H &53.81 &67.23 &67.22 &64.61 &72.25 &67.43 &73.25 &72.24 &75.07 &\textbf{76.15} & {\color{darkgreen}+1.08} \\
         \midrule

         &Base &40.88 &82.25 &77.88 &78.03 &80.28 &82.13 &\textbf{82.78} &81.96 &82.42 &82.62 & {\color{red}-0.16} \\
         &New  &96.18 &94.51 &94.84 &95.05 &95.89 &94.98 &94.56 &94.96 &\textbf{96.84} &97.29 & {\color{darkgreen}+0.45} \\
         \multirow{-3}{*}{\textbf{BTMRI}}&H &57.37 &87.95 &85.53 &85.69 &87.39 &88.09 &88.28 &87.98 &89.05 &\textbf{89.36} & {\color{darkgreen}+0.31} \\
         \midrule

         &Base &53.96 &75.92 &77.28 &75.42 &76.21 &75.19 &77.49 &75.50 &75.91 &\textbf{78.19} & {\color{darkgreen}+0.70} \\
         &New  &89.43 &90.07 &87.61 &89.61 &87.10 &90.34 &85.24 &90.37 &91.63 &89.80 & {\color{red}-1.83} \\
         \multirow{-3}{*}{\textbf{COVID-QU-Ex}}&H &67.31 &82.39 &82.12 &81.90 &81.29 &82.07 &81.18 &82.27 &83.03 &\textbf{83.59} & {\color{darkgreen}+0.56} \\
         \midrule

         &Base &38.55 &82.24 &81.96 &81.67 &85.24 &83.86 &\textbf{87.03} &84.69 &86.93 &86.13 & {\color{red}-0.90} \\ 
         &New  &52.99 &67.92 &56.56 &58.45 &76.27 &63.01 &74.82 &75.20 &\textbf{78.94} &76.02 & {\color{red}-2.92} \\
         \multirow{-3}{*}{\textbf{CTKIDNEY}}&H &44.63 &74.40 &66.93 &68.14 &80.51 &71.96 &80.46 &79.66 &\textbf{82.74} &80.76 & {\color{red}-1.98} \\
         \midrule

         &Base &34.95 &48.06 &42.88 &36.41 &40.52 &35.52 &45.76 &43.11 &54.86 &\textbf{55.83} & {\color{darkgreen}+0.97} \\
         &New  &49.59 &59.41 &60.66 &47.31 &68.13 &63.28 &72.30 &69.38 &74.10 &\textbf{76.08} & {\color{darkgreen}+1.98} \\
         \multirow{-3}{*}{\textbf{DermaMNIST}}&H &41.00 &53.14 &50.24 &41.15 &50.82 &45.50 &56.05 &53.18 &63.04 &\textbf{64.40} & {\color{darkgreen}+1.36} \\
         \midrule

         &Base &75.00 &86.22 &85.94 &81.56 &83.28 &82.89 &84.59 &85.00 &\textbf{86.50} &86.17 & {\color{red}-0.33} \\
         &New  &60.50 &58.06 &53.95 &59.00 &62.78 &60.45 &60.14 &\textbf{61.85} &61.83 &61.67 & {\color{red}-0.18} \\
         \multirow{-3}{*}{\textbf{Kvasir}}&H &66.97 &69.39 &66.29 &68.47 &71.59 &69.91 &70.30 &71.60 &\textbf{72.11} &71.89 & {\color{red}-0.22} \\
         \midrule

         &Base &37.63 &89.41 &87.77 &75.45 &90.26 &82.98 &88.12 &91.34 &88.87 &\textbf{92.82} & {\color{darkgreen}+1.48} \\
         &New  &40.69 &35.11 &42.51 &38.70 &46.14 &44.19 &50.62 &48.50 &42.73 &\textbf{62.10} & {\color{darkgreen}+11.48} \\
         \multirow{-3}{*}{\textbf{CHMNIST}}&H &39.10 &50.42 &57.28 &51.16 &61.06 &57.67 &64.30 &63.36 &57.71 &\textbf{74.41} & {\color{darkgreen}+10.11} \\
         \midrule

         &Base &59.73 &90.12 &88.33 &88.13 &90.31 &90.29 &92.78 &88.42 &\textbf{93.77} &92.00 & {\color{red}-1.77} \\
         &New  &87.60 &87.55 &95.02 &86.44 &90.16 &85.47 &94.16 &85.21 &97.00 &\textbf{98.80} & {\color{darkgreen}+1.80} \\
         \multirow{-3}{*}{\textbf{LC25000}}&H &71.03 &88.82 &91.55 &87.28 &90.23 &87.81 &93.46 &86.79 &\textbf{95.36} &95.28 & {\color{red}-0.08} \\
         \midrule

         &Base &45.18 &70.98 &66.88 &60.77 &66.16 &68.77 &69.37 &64.75 &68.46 &\textbf{72.99} & {\color{darkgreen}+2.01} \\
         &New  &55.28 &56.90 &65.56 &54.91 &62.48 &58.43 &56.95 &60.38 &\textbf{67.72} &59.06 & {\color{red}-8.66} \\
         \multirow{-3}{*}{\textbf{RETINA}}&H &49.72 &63.16 &66.21 &57.69 &64.27 &63.18 &62.55 &62.49 &\textbf{68.09} &65.29 & {\color{red}-2.80} \\
         \midrule

         &Base &35.89 &38.28 &34.08 &37.94 &39.58 &40.88 &42.58 &43.19 &44.23 &\textbf{44.93} & {\color{darkgreen}+0.70} \\
         &New  &71.90 &47.69 &63.14 &61.19 &74.10 &59.12 &76.29 &72.49 &78.35 &\textbf{80.66} & {\color{darkgreen}+2.31} \\
         \multirow{-3}{*}{\textbf{KneeXray}}&H &47.88 &42.47 &44.27 &46.84 &51.60 &48.34 &54.66 &54.13 &56.54 &\textbf{57.71} & {\color{darkgreen}+1.17} \\
         \midrule

         &Base &56.60 &75.00 &79.60 &68.20 &78.66 &74.20 &\textbf{82.33} &80.20 &80.33 &80.20 & {\color{red}-2.13} \\
         &New  &50.00 &50.23 &50.47 &50.13 &\textbf{51.59} &50.02 &48.17 &49.02 &50.07 &50.00 & {\color{red}-1.59} \\
         \multirow{-3}{*}{\textbf{OCTMNIST}}&H &53.10 &60.17 &61.77 &57.79 &\textbf{62.31} &59.76 &60.78 &60.85 &61.69 &61.60 & {\color{red}-0.71} \\
         
    \bottomrule
    \end{tabular}}
    \label{tab:b2nbiomed}
\end{table*}

\begin{table*}[!ht]
    \centering
    \caption{\textbf{Calibration performance using the ECE metric, on few-shot learning task across 11 biomedical datasets.}}
    \vspace{-0.2cm}
    \scalebox{0.67}{
    \begin{tabular}{l|ccccccccc|c}
    \toprule
         &\multicolumn{1}{c}{BioMedCLIP} &\multicolumn{1}{c}{CoOp} &\multicolumn{1}{c}{CoCoOp} &\multicolumn{1}{c}{KgCoOp} &\multicolumn{1}{c}{MaPLe} &\multicolumn{1}{c}{ProGrad} &\multicolumn{1}{c}{PromptSRC} &\multicolumn{1}{c}{TCP} &\multicolumn{1}{c}{BiomedCoOp}& BioVLM \\
        
        \multirow{-2}{*}{\textbf{Source}} &NEJM AI'25 &IJCV'22 &CVPR'22 &CVPR'23 &CVPR'23 &ICCV'23 &ICCV'23 &CVPR'24 &CVPR'25 & (Ours) \\ 
        \midrule

        \textbf{BTMRI}	&0.17	&0.11	&0.13	&0.11	&0.08	&0.12	&0.09	&0.08	&0.09	&\textbf{0.06} \\
\textbf{BUSI}	&0.20	&0.12	&0.15	&0.11	&0.13	&0.15	&0.14	&0.13	&0.12	&\textbf{0.10} \\
\textbf{COVID-QU-Ex}	&0.08	&0.05	&0.10	&0.04	&0.03	&0.06	&0.07	&0.04	&0.03	&\textbf{0.01} \\
\textbf{CTKIDNEY}	&0.22	&0.12	&0.20	&0.11	&0.11	&0.14	&0.13	&0.15	&0.13	&\textbf{0.09} \\
\textbf{DermaMNIST}	&0.14	&0.11	&0.16	&0.10	&0.09	&0.14	&0.10	&0.09	&0.10	&\textbf{0.07} \\
\textbf{Kvasir}	&0.10	&0.09	&0.11	&0.08	&\textbf{0.05}	&0.09	&0.07	&0.08	&0.08	&0.06 \\
\textbf{CHMNIST}	&0.08	&0.06	&0.06	&0.05	&0.05	&0.08	&0.09	&0.05	&0.04	&\textbf{0.03} \\
\textbf{LC25000}	&0.07	&0.05	&0.06	&0.06	&\textbf{0.03}	&0.09	&0.05	&0.06	&0.05	&0.04 \\
\textbf{RETINA}	&0.12	&0.10	&0.11	&0.10	&0.07	&0.10	&0.08	&0.07	&0.07	&\textbf{0.05} \\
\textbf{KneeXray}	&0.09	&0.05	&0.06	&0.05	&0.04	&0.08	&0.06	&0.06	&0.06	&\textbf{0.02} \\
\textbf{OCTMNIST}	&0.16	&0.11	&0.11	&0.10	&0.11	&0.13	&0.10	&0.11	&0.14	&\textbf{0.09} \\
    \bottomrule
        
    \end{tabular}}
    \vspace{-0.2cm}
    \label{tab:calibration}
\end{table*}

\begin{table*}[!ht]
    \centering
    \caption{\textbf{Comparison of PEFT methods on the few-shot learning task of MedMNIST+ benchmark.}}
    \vspace{-0.2cm}
    \scalebox{0.7}{
    \begin{tabular}{l|cccccc|c}
    \toprule
         \multirow{-1}{*}{\textbf{Dataset}} &\multicolumn{1}{c}{LoRA} &\multicolumn{1}{c}{AdaLoRA}  &\multicolumn{1}{c}{LayerNorm} &\multicolumn{1}{c}{BitFit} &\multicolumn{1}{c}{CLIP-Adapter} &\multicolumn{1}{c|}{Tip-Adapter} & BioVLM (Ours) \\

        \midrule

        \textbf{PathMNIST}	&74.73	&75.29	&61.39	&59.28	&71.47	&72.59	&\cellcolor[gray]{0.9}\textbf{81.56} \\   
         
        \textbf{DermaMNIST}	&36.18	&37.50	&34.48	&32.52	&37.29	&38.11	&\cellcolor[gray]{0.9}\textbf{45.27} \\
        
        \textbf{OCTMNIST}	&59.19	&59.91	&48.28	&44.59	&58.12	&57.30	&\cellcolor[gray]{0.9}\textbf{62.57} \\
        
        \textbf{PneumoniaMNIST}	&70.36	&72.72	&58.42	&52.96	&68.40	&69.24	&\cellcolor[gray]{0.9}\textbf{79.65} \\
        
        \textbf{RetinaMNIST}	&35.49	&36.81	&30.58	&27.32	&34.67	&33.78	&\cellcolor[gray]{0.9}\textbf{40.08} \\
        
        \textbf{BreastMNIST} &58.50	&57.28	&47.38	&45.10	&56.39	&57.00	&\cellcolor[gray]{0.9}\textbf{65.60} \\
        
        \textbf{BloodMNIST}	&45.28	&45.43	&25.28	&22.49	&40.42	&41.58	&\cellcolor[gray]{0.9}\textbf{70.15} \\
        
        \textbf{TissueMNIST} &17.39	&18.06	&16.47	&15.59	&15.93	&16.78	&\cellcolor[gray]{0.9}\textbf{27.34} \\
        
        \textbf{OrganAMNIST}	&36.61	&37.06	&34.08	&32.81	&36.51	&36.90	&\cellcolor[gray]{0.9}\textbf{54.75} \\
        
        \textbf{OrganCMNIST}	&35.40	&37.47	&29.48	&26.33	&35.02	&34.85	&\cellcolor[gray]{0.9}\textbf{49.22} \\
        
        \textbf{OrganSMNIST}	&33.95	&35.19	&28.70	&25.60	&32.76	&33.04	&\cellcolor[gray]{0.9}\textbf{45.41} \\
        
       \midrule
       \textbf{Average} &45.73	&46.61	&37.69	&34.96	&44.27	&44.65	&\cellcolor[gray]{0.9}\textbf{56.51} \\
    
    \bottomrule
        
    \end{tabular}}
    \vspace{-0.2cm}
    \label{tab:peft}
\end{table*}

\begin{table*}[!ht]
    \centering
    \caption{\textbf{OOD generalization, with source dataset as PathMNIST.}}
    \vspace{-0.2cm}
    \scalebox{0.82}{
    \begin{tabular}{l|cccccccccc|c}
    \toprule
        Method &Derma &OCT &Pneumonia  &Retina &Breast &Blood &Tissue &OrganA &OrganC &OrganS &Average \\
        \midrule

        CoOp &\textbf{66.90} &31.80 &65.97 &19.42 &66.67 &12.86 &3.93 &18.77 &16.11 &16.96 &31.94 \\
        CoCoOp &66.25 &\textbf{34.40} &62.50 &16.50 &50.85 &16.92 &3.89 &16.88 &14.05 &14.53 &29.68 \\
        KgCoOp &66.81 &30.90 &66.77 &18.75 &\textbf{67.09} &16.28 &4.33 &19.86 &16.62 &16.88 &32.43 \\
        {MaPLe} &66.51 &29.85 &66.76 &20.50 &64.93 &\textbf{19.04} &4.27 &19.34 &16.90 &17.04 &32.51 \\
        ProGrad &66.76 &32.13 &68.75 &\textbf{23.17} &48.72 &18.28 &5.53 &21.45 &17.55 &\textbf{18.47} &32.08 \\
        {PromptSRC} &66.73 &31.45 &65.23 &18.47 &62.49 &17.38 &\textbf{5.83} &18.76 &17.25 &17.46 &32.11 \\
        {TCP} &66.34 &30.78 &69.45 &17.24 &61.82 &17.65 &4.05 &19.78 &17.35 &17.32 &32.18 \\
        BioMedCoOp &66.85 &33.27 &\textbf{71.69} &17.33 &61.75 &18.99 &4.41 &19.36 &16.33 &17.26 &\textbf{32.72} \\
        \midrule
        \cellcolor[gray]{0.9}BioVLM (Ours) &\cellcolor[gray]{0.9}66.52 &\cellcolor[gray]{0.9}26.33 &\cellcolor[gray]{0.9}62.23 &\cellcolor[gray]{0.9}17.75 &\cellcolor[gray]{0.9}61.32 &\cellcolor[gray]{0.9}16.83 &\cellcolor[gray]{0.9}4.55 &\cellcolor[gray]{0.9}\textbf{23.20} &\cellcolor[gray]{0.9}\textbf{18.42} &\cellcolor[gray]{0.9}18.32 &\cellcolor[gray]{0.9}31.55 \\
    
    \bottomrule
    \end{tabular}}
    \vspace{-0.3cm}
    \label{tab:fewshot_1}
\end{table*}

\begin{table*}[!ht]
    \centering
    \caption{\textbf{OOD generalization, with source dataset as DermaMNIST.}}
    \vspace{-0.2cm}
    \scalebox{0.82}{
    \begin{tabular}{l|cccccccccc|c}
    \toprule
        Method &Path &OCT &Pneumonia &Retina &Breast &Blood &Tissue &OrganA &OrganC &OrganS &Average \\
        \midrule

        CoOp &16.72 &25.33 &54.54 &23.50 &32.27 &16.91 &\textbf{4.39} &17.92 &17.38 &18.36 &22.73 \\
        CoCoOp &17.70 &\textbf{38.50} &65.33 &28.42 &\textbf{47.87} &19.76 &3.83 &14.33 &14.61 &16.37 &26.67 \\
        KgCoOp &15.46 &25.27 &55.29 &33.67 &29.28 &16.91 &4.35 &18.24 &16.91 &17.56 &23.29 \\
        {MaPLe} &20.56	&26.43	&55.21	&30.53	&30.26	&12.78	&2.65	&15.79	&16.72	&16.98	&22.79 \\
        ProGrad &16.94 &25.03 &56.04 &25.67 &29.91 &11.32 &3.80 &19.24 &17.83 &19.36 &22.51 \\
        {PromptSRC} &24.87	&28.42	&59.23	&35.61	&25.47	&16.21	&2.57	&18.94	&17.33	&15.53	&24.42 \\
        {TCP} &16.42	&25.89	&60.31	&26.78	&32.47	&13.45	&2.99	&16.43	&17.31	&19.67	&23.17 \\
        BioMedCoOp &29.08 &31.93 &\textbf{66.45} &43.50 &29.91 &\textbf{23.64} &3.78 &21.73 &\textbf{20.59} &\textbf{23.16} &29.38 \\
        \midrule
        \cellcolor[gray]{0.9}BioVLM (Ours) &\cellcolor[gray]{0.9}\textbf{34.25} &\cellcolor[gray]{0.9}27.87 &\cellcolor[gray]{0.9}62.45 &\cellcolor[gray]{0.9}\textbf{43.67} &\cellcolor[gray]{0.9}41.24 &\cellcolor[gray]{0.9}17.64 &\cellcolor[gray]{0.9}3.72 &\cellcolor[gray]{0.9}\textbf{23.08} &\cellcolor[gray]{0.9}20.20 &\cellcolor[gray]{0.9}19.98 &\cellcolor[gray]{0.9}\textbf{29.41} \\
    
    \bottomrule
    \end{tabular}}
    \vspace{-0.3cm}
    \label{tab:ood_no_derma}
\end{table*}

\begin{table*}[!ht]
    \centering
    \caption{\textbf{OOD generalization, with source dataset as OCTMNIST.}}
    \vspace{-0.2cm}
    \scalebox{0.82}{
    \begin{tabular}{l|cccccccccc|c}
    \toprule
        Method &Path &Derma &Pneumonia &Retina &Breast &Blood &Tissue &OrganA &OrganC &OrganS &Average \\
        \midrule

        CoOp &41.91 &64.24 &62.50 &\textbf{37.25} &30.34 &18.44 &3.64 &19.00 &17.77 &20.50 &31.56 \\
        CoCoOp &21.06 &55.68 &62.82 &35.67 &\textbf{35.90} &16.93   &\textbf{6.20} &20.50 &18.56 &18.38 &29.17 \\
        KgCoOp &41.62 &62.34 &62.50 &35.58 &30.98 &18.69 &3.71 &19.84 &18.66 &21.09 &31.50 \\
        {MaPLe} &42.12	&62.78	&62.53	&33.67	&31.37	&\textbf{20.24}	&4.78	&19.45	&20.13	&22.46	&31.95 \\
        ProGrad &38.90 &62.69 &62.50 &35.50 &29.70 &17.21 &4.14 &20.12 &18.36 &21.16 &31.03 \\
       {PromptSRC} &40.56	&61.37	&\textbf{63.42}	&35.91	&32.55	&18.43	&5.84	&23.65	&21.18	&21.39	&32.43 \\
        {TCP} &40.96	&61.02	&62.15	&36.24	&34.98	&19.54	&6.32	&21.52	&21.05	&20.87	&32.47 \\
        BioMedCoOp &37.98 &61.88 &62.50 &33.58 &29.70 &19.65 &4.54 &21.49 &20.24 &22.94 &31.45 \\
        \midrule
        \cellcolor[gray]{0.9}BioVLM (Ours) &\cellcolor[gray]{0.9}\textbf{44.62} &\cellcolor[gray]{0.9}\textbf{66.05} &\cellcolor[gray]{0.9}62.50 &\cellcolor[gray]{0.9}30.25 &\cellcolor[gray]{0.9}30.34 &\cellcolor[gray]{0.9}16.87 &\cellcolor[gray]{0.9}4.46 &\cellcolor[gray]{0.9}\textbf{28.00} &\cellcolor[gray]{0.9}\textbf{22.32} &\cellcolor[gray]{0.9}\textbf{23.46} &\cellcolor[gray]{0.9}\textbf{32.89} \\
    
    \bottomrule
    \end{tabular}}
    \vspace{-0.3cm}
    \label{tab:ood_no_oct}
\end{table*}

\begin{table*}[!ht]
    \centering
    \caption{\textbf{OOD generalization, with source dataset as PneumoniaMNIST.}}
    \vspace{-0.2cm}
    \scalebox{0.82}{
    \begin{tabular}{l|cccccccccc|c}
    \toprule
        Method &Path &Derma &OCT &Retina &Breast &Blood &Tissue &OrganA &OrganC &OrganS &Average \\
        \midrule

        CoOp &17.13 &40.00 &25.73 &24.33 &28.85 &16.91 &\textbf{5.58} &19.62 &16.86 &16.78 &21.18 \\
        CoCoOp &17.87 &54.50 &31.37 &37.25 &\textbf{43.78} &16.50 &4.15 &21.77 &15.85 &17.49 &26.05 \\
        KgCoOp &17.88 &43.76 &26.27 &25.83 &27.78 &16.91 &5.55 &21.09 &17.87 &18.26 &22.12 \\
        {MaPLe} &22.32	&56.93	&31.45	&39.67	&27.89	&16.37	&4.87	&20.64	&16.34	&17.38	&25.39 \\
        ProGrad &17.89 &52.43 &27.60 &32.83 &27.56 &16.91 &5.48 &19.02 &16.67 &17.04 &23.34 \\
        {PromptSRC} &20.56	&52.46	&32.07	&36.68	&31.45	&16.06	&4.93	&21.05	&16.89	&17.93	&25.01 \\
        {TCP} &24.78	&57.25	&30.35	&40.29	&34.67	&16.14	&4.37	&19.53	&16.23	&17.85	&26.15 \\
        BioMedCoOp &26.38 &55.31 &\textbf{33.87} &42.00 &35.90 &\textbf{17.00} &4.91 &17.09 &16.69 &\textbf{19.80} &26.89 \\
        \midrule
        \cellcolor[gray]{0.9}BioVLM (Ours) &\cellcolor[gray]{0.9}\textbf{30.14} &\cellcolor[gray]{0.9}\textbf{63.66} &\cellcolor[gray]{0.9}33.20 &\cellcolor[gray]{0.9}\textbf{44.25} &\cellcolor[gray]{0.9}30.56 &\cellcolor[gray]{0.9}16.87 &\cellcolor[gray]{0.9}3.85 &\cellcolor[gray]{0.9}\textbf{22.09} &\cellcolor[gray]{0.9}\textbf{18.48} &\cellcolor[gray]{0.9}18.76 &\cellcolor[gray]{0.9}\textbf{28.19} \\
    
    \bottomrule
    \end{tabular}}
    \vspace{-0.3cm}
    \label{tab:ood_no_pneumonia}
\end{table*}

\begin{table*}[!ht]
    \centering
    \caption{\textbf{OOD generalization, with source dataset as RetinaMNIST.}}
    \vspace{-0.2cm}
    \scalebox{0.82}{
    \begin{tabular}{l|cccccccccc|c}
    \toprule
    Method &Path &Derma &OCT &Pneumonia &Breast &Blood &Tissue &OrganA &OrganC &OrganS &Average \\
    \midrule
    CoOp        &\textbf{41.93} &64.35 &45.73 &64.10 &\textbf{45.67} &19.71 &6.85 &20.18 &17.94 &20.37 &\textbf{34.68} \\
    CoCoOp      &26.54 &58.72 &32.27 &61.91 &42.73 &14.31 &3.69 &\textbf{23.24} &18.67 &18.50 &30.06 \\
    KgCoOp      &40.43 &63.70 &45.93 &63.94 &35.90 &18.48 &6.84 &19.76 &16.98 &19.81 &33.18 \\
    {MaPLe} &38.53	&62.47	&42.66	&62.40	&33.56	&14.68	&4.09	&19.47	&16.90	&\textbf{22.05}	&31.68 \\
    ProGrad     &38.78 &63.89 &43.80 &\textbf{65.33} &39.31 &17.71 &\textbf{6.95} &20.15 &19.11 &21.48 &33.65 \\
    {PromptSRC} &35.69	&62.89	&45.15	&63.85	&37.64	&16.43	&4.78	&18.28	&17.28	&20.38	&32.24 \\
    {TCP} &34.65	&62.50	&46.34	&64.16	&32.59	&16.78	&4.15	&19.78	&20.67	&18.30	&31.99 \\
    BioMedCoOp  &34.69 &\textbf{64.66} &\textbf{48.63} &62.50 &31.84 &18.56 &5.64 &20.89 &\textbf{20.05} &21.76 &32.92 \\
    \midrule
    \cellcolor[gray]{0.9}BioVLM (Ours)&\cellcolor[gray]{0.9}37.57 &\cellcolor[gray]{0.9}63.61 &\cellcolor[gray]{0.9}32.90 &\cellcolor[gray]{0.9}62.34 &\cellcolor[gray]{0.9}34.83 &\cellcolor[gray]{0.9}\textbf{21.88} &\cellcolor[gray]{0.9}4.96 &\cellcolor[gray]{0.9}20.54 &\cellcolor[gray]{0.9}16.63 &\cellcolor[gray]{0.9}18.46 &\cellcolor[gray]{0.9}31.37 \\
    \bottomrule
    \end{tabular}
    }
    \vspace{-0.3cm}
    \label{tab:ood_no_retina}
\end{table*}

\begin{table*}[!ht]
    \centering
    \caption{\textbf{OOD generalization, with source dataset as BreastMNIST.}}
    \vspace{-0.2cm}
    \scalebox{0.82}{
    \begin{tabular}{l|cccccccccc|c}
    \toprule
    Method &Path &Derma &OCT &Pneumonia &Retina &Blood &Tissue &OrganA &OrganC &OrganS &Average \\
    \midrule
    CoOp         &32.08 &\textbf{65.68} &34.33 &74.62 &36.83 &\textbf{19.08} &4.06 &22.62 &19.40 &20.37 &32.91 \\
    CoCoOp       &26.15 &51.89 &32.97 &63.03 &28.83 &16.56 &3.59 &20.72 &18.62 &19.40 &28.18 \\
    KgCoOp       &36.46 &63.09 &29.73 &\textbf{75.38} &41.42 &17.90 &3.46 &21.75 &20.35 &21.61 &\textbf{33.12} \\
    {MaPLe} &33.54	&50.42	&30.52	&60.17	&40.23	&16.49	&3.58	&24.79	&20.47	&21.35	&30.16 \\
    ProGrad      &33.86 &46.04 &\textbf{37.63} &72.60 &32.17 &17.26 &\textbf{5.76} &24.68 &\textbf{23.69} &23.59 &31.73 \\
    {PromptSRC} &34.58	&55.18	&34.11	&63.35	&36.14	&18.32	&3.16	&25.68	&19.13	&20.55	&31.02 \\
    {TCP} &33.16	&52.90	&31.38 &67.49	&39.87	&17.03	&3.96	&25.31	&17.43	&18.94	&30.75 \\
    BioMedCoOp   &31.76 &45.47 &33.60 &66.78 &\textbf{44.08} &17.07 &3.46 &26.46 &23.65 &\textbf{26.29} &31.86 \\
    \midrule
    \cellcolor[gray]{0.9}BioVLM (Ours)&\cellcolor[gray]{0.9}\textbf{41.52} &\cellcolor[gray]{0.9}65.20 &\cellcolor[gray]{0.9}32.60 &\cellcolor[gray]{0.9}62.39 &\cellcolor[gray]{0.9}37.08 &\cellcolor[gray]{0.9}16.92 &\cellcolor[gray]{0.9}4.19 &\cellcolor[gray]{0.9}\textbf{26.69} &\cellcolor[gray]{0.9}19.37 &\cellcolor[gray]{0.9}19.62 &\cellcolor[gray]{0.9}32.56 \\
    \bottomrule
    \end{tabular}}
    \vspace{-0.3cm}
    \label{tab:ood_nobreast}
\end{table*}

\begin{table*}[!ht]
    \centering
    \caption{\textbf{OOD generalization, with source dataset as BloodMNIST.}}
    \vspace{-0.2cm}
    \scalebox{0.82}{
    \begin{tabular}{l|cccccccccc|c}
    \toprule
    Method &Path &Derma &OCT &Pneumonia &Retina &Breast &Tissue &OrganA &OrganC &OrganS &Average \\
    \midrule
    CoOp         &\textbf{35.04} &65.10 &26.60 &67.63 &23.98 &52.14 &5.18 &19.63 &17.18 &18.74 &33.12 \\
    CoCoOp       &26.50 &63.03 &29.20 &66.93 &24.00 &\textbf{57.48} &\textbf{5.29} &17.24 &15.02 &14.40 &31.91 \\
    KgCoOp       &30.78 &63.49 &26.80 &66.99 &22.58 &48.51 &4.94 &20.27 &17.17 &18.51 &32.00 \\
    {MaPLe} &30.67	&65.38	&28.04	&63.59	&20.14 &45.09	&4.65	&18.43	&18.32	&18.84	&31.32 \\
    ProGrad      &29.08 &66.83 &25.43 &68.64 &29.42 &42.74 &5.08 &20.47 &\textbf{18.43} &20.27 &32.64 \\
    {PromptSRC} &33.14	&65.82	&28.59	&64.72	&24.64	&47.30	&4.87	&18.94	&17.45	&\textbf{20.56}	&32.60 \\
    {TCP} &31.08	&64.34	&30.15	&61.46	&22.60	&48.85	&4.96	&19.07	&16.95	&18.54	&31.80 \\
    BioMedCoOp   &32.67 &\textbf{66.86} &29.00 &\textbf{68.91} &21.67 &42.73 &\textbf{5.29} &18.69 &16.41 &18.48 &32.07 \\
    \midrule
    \cellcolor[gray]{0.9}BioVLM (Ours)&\cellcolor[gray]{0.9}29.30 &\cellcolor[gray]{0.9}66.66 &\cellcolor[gray]{0.9}\textbf{34.77} &\cellcolor[gray]{0.9}62.71 &\cellcolor[gray]{0.9}\textbf{30.75} &\cellcolor[gray]{0.9}50.64 &\cellcolor[gray]{0.9}5.04 &\cellcolor[gray]{0.9}\textbf{23.27} &\cellcolor[gray]{0.9}17.77 &\cellcolor[gray]{0.9}19.17 &\cellcolor[gray]{0.9}\textbf{34.01} \\
    \bottomrule
    \end{tabular}}
    \vspace{-0.3cm}
    \label{tab:ood_noblood}
\end{table*}

\begin{table*}[!ht]
    \centering
    \caption{\textbf{OOD generalization, with source dataset as TissueMNIST.}}
    \vspace{-0.2cm}
    \scalebox{0.82}{
    \begin{tabular}{l|cccccccccc|c}
    \toprule
    Method &Path &Derma &OCT &Pneumonia &Retina &Breast &Blood &OrganA &OrganC &OrganS &Average \\
    \midrule
    CoOp         &18.67 &53.05 &28.40 &52.46 &29.92 &\textbf{64.74} &\textbf{16.92} &18.00 &14.44 &14.03 &31.06 \\
    CoCoOp       &32.66 &49.34 &27.77 &61.32 &\textbf{43.50} &34.19 &16.91 &13.75 &13.58 &14.26 &30.73 \\
    KgCoOp       &27.84 &65.65 &33.57 &62.66 &33.58 &62.39 &16.90 &17.38 &14.00 &14.82 &34.88 \\
    {MaPLe} &37.58	&65.31	&29.31	&61.39	&27.48	&39.47	&16.90	&18.05	&16.58	&17.48	&32.96 \\
    ProGrad      &38.26 &65.83 &25.90 &62.61 &25.33 &36.96 &16.90 &\textbf{25.66} &\textbf{19.94} &18.63 &33.60 \\
    {PromptSRC} &40.25	&64.64	&30.54	&62.58	&27.92	&42.48	&16.89	&21.36	&16.84	&17.85	&34.14 \\
    {TCP} &38.54	&64.92	&30.26	&62.80	&28.03	&30.42	&16.90	&20.58	&17.07	&16.30	&32.58 \\
    BioMedCoOp   &40.18 &65.98 &\textbf{44.67} &\textbf{64.74} &27.25 &37.82 &16.90 &19.66 &15.33 &16.59 &34.91 \\
    \midrule
    \cellcolor[gray]{0.9}BioVLM (Ours)&\cellcolor[gray]{0.9}\textbf{48.29} &\cellcolor[gray]{0.9}\textbf{67.00} &\cellcolor[gray]{0.9}31.20 &\cellcolor[gray]{0.9}62.34 &\cellcolor[gray]{0.9}\textbf{36.83} &\cellcolor[gray]{0.9}34.83 &\cellcolor[gray]{0.9}16.90 &\cellcolor[gray]{0.9}25.05 &\cellcolor[gray]{0.9}19.63 &\cellcolor[gray]{0.9}\textbf{20.73} &\cellcolor[gray]{0.9}\textbf{36.28} \\
    \bottomrule
    \end{tabular}}
    \vspace{-0.3cm}
    \label{tab:ood_notissue}
\end{table*}

\begin{table*}[!ht]
    \centering
    \caption{\textbf{OOD generalization, with source dataset as OrganAMNIST.}}
    \vspace{-0.2cm}
    \scalebox{0.82}{
    \begin{tabular}{l|cccccccccc|c}
    \toprule
    Method & Path & Derma & OCT & Pneumonia & Retina & Breast & Blood & Tissue & OrganC & OrganS & \textbf{Average} \\
    \midrule
    CoOp         & 32.08 & 39.97 & 34.33 & \textbf{66.40} & 37.33 & 32.48 & 16.88 & 5.01 & 37.21 & 36.63 & 33.83 \\
    CoCoOp       & 26.15 & 54.68 & 32.97 & 57.69 & 35.25 & \textbf{46.37} & 16.85 & 5.10 & 36.66 & 35.23 & \textbf{34.70} \\
    KgCoOp       & \textbf{36.46} & 44.17 & 29.73 & 57.58 & 30.67 & 34.19 & \textbf{16.94} & \textbf{5.73} & 38.08 & 36.63 & 33.02 \\
    {MaPLe} &22.57	&48.42	&35.28	&60.93	&35.52	&32.64	&16.78	&5.02	&36.53	&30.68	&32.44 \\
    ProGrad      & 33.86 & \textbf{59.27} & \textbf{37.63} & 65.33 & 35.75 & 37.18 & 16.66 & 4.75 & 28.48 & 27.92 & 34.68 \\
    {PromptSRC} &25.49	&48.96	&34.50	&62.05	&36.04	&33.49	&16.92	&4.85	&35.89	&34.84	&33.30 \\
    {TCP} &26.31	&52.06	&31.06	&63.87	&33.18	&37.98	&16.88	&4.98	&36.58	&35.05	&33.80 \\
    BioMedCoOp   & 31.76 & 57.34 & 33.60 & 62.29 & 34.25 & 33.76 & 16.90 & 3.72 & 31.09 & 30.43 & 33.51 \\
    \midrule
    \cellcolor[gray]{0.9}BioVLM (Ours)& \cellcolor[gray]{0.9}21.15 & \cellcolor[gray]{0.9}51.89 & \cellcolor[gray]{0.9}30.97 & \cellcolor[gray]{0.9}62.18 & \cellcolor[gray]{0.9}\textbf{39.42} & \cellcolor[gray]{0.9}28.85 & \cellcolor[gray]{0.9}16.86 & \cellcolor[gray]{0.9}5.50 & \cellcolor[gray]{0.9}\textbf{44.51} & \cellcolor[gray]{0.9}\textbf{42.04} & \cellcolor[gray]{0.9}34.34 \\
    \bottomrule
    \end{tabular}}
    \vspace{-0.3cm}
    \label{tab:ood_no_organA}
\end{table*}

\begin{table*}[!ht]
    \centering
    \caption{\textbf{OOD generalization, with source dataset as OrganCMNIST.}}
    \vspace{-0.2cm}
    \scalebox{0.82}{
    \begin{tabular}{l|cccccccccc|c}
    \toprule
    Method & Path & Derma & OCT & Pneumonia & Retina & Breast & Blood & Tissue & OrganA & OrganS & \textbf{Average} \\
    \midrule
    CoOp         & 29.05 & 48.08 & 31.53 & 62.55 & 35.33 & 33.33 & 17.04 & 6.54 & 42.36 & 36.20 & 34.20 \\
    CoCoOp       & 21.23 & 59.68 & \textbf{37.17} & 62.34 & 40.58 & \textbf{41.67} & \textbf{17.32} & 4.09 & 37.63 & 33.02 & 35.47 \\
    KgCoOp       & 29.22 & 45.83 & 26.47 & 61.27 & 31.17 & \textbf{41.67} & 16.90 & 6.76 & 41.84 & 34.41 & 33.55 \\
    {MaPLe} &18.43	&58.73	&35.45	&62.34	&36.58	&35.78	&16.45	&5.83	&34.75	&36.59	&34.09 \\
    ProGrad      & \textbf{32.57} & \textbf{63.01} & 30.37 & \textbf{64.05} & 34.08 & 34.40 & 15.19 & \textbf{6.85} & 31.11 & 26.67 & 33.83 \\
    {PromptSRC} &20.59	&56.09	&32.07	&61.48	&37.07	&38.61	&16.70	&5.05	&36.58	&38.10	&34.23 \\
    {TCP} &22.05	&60.18	&33.86	&62.95	&36.14	&39.57	&16.36	&4.87	&40.25	&29.28	&34.55 \\
    BioMedCoOp   & 24.86 & 60.28 & 34.13 & 62.34 & 34.25 & 30.99 & 16.72 & 5.21 & 30.74 & 26.71 & 32.62 \\
    \midrule
    \cellcolor[gray]{0.9}BioVLM (Ours)& \cellcolor[gray]{0.9}19.72 & \cellcolor[gray]{0.9}62.94 & \cellcolor[gray]{0.9}30.33 & \cellcolor[gray]{0.9}62.13 & \cellcolor[gray]{0.9}\textbf{43.33} & \cellcolor[gray]{0.9}31.84 & \cellcolor[gray]{0.9}16.88 & \cellcolor[gray]{0.9}4.41 & \cellcolor[gray]{0.9}\textbf{50.40} & \cellcolor[gray]{0.9}\textbf{43.68} & \cellcolor[gray]{0.9}\textbf{36.57} \\
    \bottomrule
    \end{tabular}}
    \vspace{-0.3cm}
    \label{tab:ood_no_organc}
\end{table*}

\begin{table*}[!htbp]
    \centering
    \caption{\textbf{OOD generalization, with source dataset as OrganSMNIST.}}
    \vspace{-0.2cm}
    \scalebox{0.82}{
    \begin{tabular}{l|cccccccccc|c}
    \toprule
    Method & Path & Derma & OCT & Pneumonia & Retina & Breast & Blood & Tissue & OrganA & OrganC & \textbf{Average} \\
    \midrule
    CoOp         & 33.43 & 62.69 & 27.27 & 62.29 & 32.50 & 31.84 & \textbf{17.01} & \textbf{5.42} & 39.08 & 36.25 & 34.78 \\
    CoCoOp       & 19.63 & 43.44 & 29.17 & 61.75 & 34.00 & \textbf{46.36} & 16.52 & 4.56 & 34.99 & 33.39 & 32.38 \\
    KgCoOp       & 31.80 & 57.12 & 26.77 & 61.86 & 34.25 & 39.75 & 16.94 & 5.35 & 39.32 & 37.25 & 35.04 \\
    {MaPLe} &20.55	&62.68	&28.95	&61.58	&33.28	&32.57	&16.70	&4.27	&36.44	&36.27	&33.33 \\
    ProGrad      & \textbf{34.58} & \textbf{65.72} & 26.33 & \textbf{68.27} & 34.17 & 33.76 & 16.64 & 4.62 & 30.96 & 28.10 & 34.32 \\
    {PromptSRC} &22.97	&64.00	&27.40	&63.20	&30.57	&30.18	&16.54	&4.65	&35.98	&36.90	&33.24 \\
    {TCP} &25.48	&61.49	&23.56	&62.45	&35.19	&33.79	&16.49	&4.50	&33.29	&30.51	&32.68 \\
    BioMedCoOp   & 28.65 & 62.14 & \textbf{35.83} & 62.39 & 33.67 & 31.62 & 16.77 & 4.11 & 31.94 & 30.29 & 33.74 \\
    \midrule
    \cellcolor[gray]{0.9}BioVLM (Ours)& \cellcolor[gray]{0.9}22.38 &\cellcolor[gray]{0.9}63.19 &\cellcolor[gray]{0.9} 26.50 & \cellcolor[gray]{0.9}62.23 & \cellcolor[gray]{0.9}\textbf{41.08} & \cellcolor[gray]{0.9}37.18 & \cellcolor[gray]{0.9}16.90 & \cellcolor[gray]{0.9}4.40 & \cellcolor[gray]{0.9}\textbf{41.91} & \cellcolor[gray]{0.9}\textbf{42.51} & \cellcolor[gray]{0.9}\textbf{35.83} \\
    \bottomrule
    \end{tabular}}
    \vspace{-0.3cm}
    \label{tab:ood_no_organs}
\end{table*}

\end{document}